\documentclass[10pt,twocolumn,letterpaper]{article}

\usepackage{cvpr}              

\usepackage{algorithm}
\usepackage{algpseudocode}
\usepackage{multirow}
\usepackage{booktabs}
\usepackage{adjustbox}

\usepackage{color}
\usepackage{colortbl}
\usepackage{xcolor}

\usepackage{array}
\usepackage{multirow}
\usepackage{makecell}

\definecolor{lightyellow}{RGB}{255,254,242}

\definecolor{cvprblue}{rgb}{0.21,0.49,0.74}
\usepackage[pagebackref,breaklinks,colorlinks,allcolors=cvprblue]{hyperref}

\def\paperID{3260} 
\def\confName{CVPR}
\def\confYear{2025}

\title{UADet: A Remarkably Simple Yet Effective Uncertainty-Aware Open-Set Object Detection Framework}
\author{Silin Cheng \qquad\qquad Yuanpei Liu \qquad\qquad Kai Han\thanks{Corresponding author.}\\
Visual AI Lab, The University of Hong Kong\\
{\tt\small hnslcheng@connect.hku.hk\qquad  ypliu0@connect.hku.hk\qquad kaihanx@hku.hk}
}

\begin{document}
\maketitle
\begin{abstract}
We tackle the challenging problem of Open-Set Object Detection (OSOD), which aims to detect both known and unknown objects in unlabelled images. The main difficulty arises from the absence of supervision for these unknown classes, making it challenging to distinguish them from the background. Existing OSOD detectors either fail to properly exploit or inadequately leverage the abundant unlabeled unknown objects in training data, restricting their performance. To address these limitations, we propose UADet, an Uncertainty-Aware Open-Set Object Detector that considers appearance and geometric uncertainty. By integrating these uncertainty measures, UADet effectively reduces the number of unannotated instances incorrectly utilized or omitted by previous methods. Extensive experiments on OSOD benchmarks demonstrate that UADet substantially outperforms previous state-of-the-art (SOTA) methods in detecting both known and unknown objects, achieving a 1.8× improvement in unknown recall while maintaining high performance on known classes. When extended to Open World Object Detection (OWOD), our method shows significant advantages over the current SOTA method, with average improvements of 13.8\% and 6.9\% in unknown recall on M-OWODB and S-OWODB benchmarks, respectively. Extensive results validate the effectiveness of our uncertainty-aware approach across different open-set scenarios.

\end{abstract}
\section{Introduction}
\label{sec:intro}

\begin{figure}[t]
    \centering
    \includegraphics[width=1.0\linewidth]{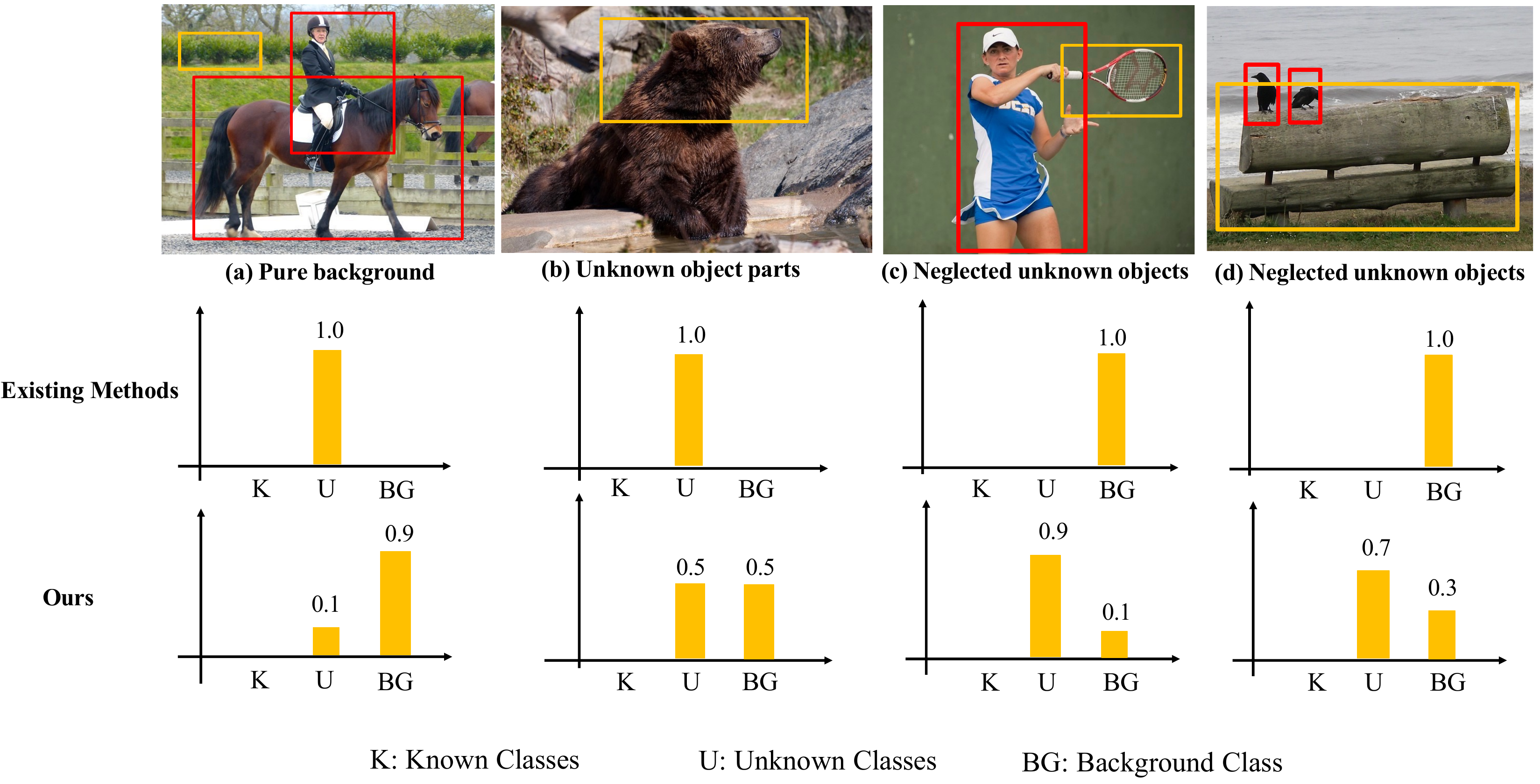}
    \caption{\textbf{A comparison of the proposed uncertainty-aware labeling strategy (bottom) with previous methods which typically assign hard labels (top) for unknown objects.} The \textcolor{red}{red} boxes depict ground-truth boxes, while the \textcolor{orange}{yellow} boxes show proposals that do not match the ground-truth boxes. The predicted scores for each negative proposal are also displayed. 
    \textbf{(a) Pure background:} Negative proposals that correspond to the real background are misclassified as an unknown class. 
    \textbf{(b) Unknown object parts:} Negative proposals representing partial parts of unknown objects are overconfidently classified as the unknown class.
    \textbf{(c)(d) Neglected unknown objects:} Negative proposals are misclassified as ``background", resulting in the omission of unknown objects. In contrast, our proposed pseudo label strategy can assign more appropriate pseudo labels for the above cases.}
    \label{fig:intro}
    \vspace{-3mm}
\end{figure}

Object detection is a fundamental and challenging problem in computer vision, aiming to accurately localize and recognize objects of specific categories in images. Over the past decades, the field of object detection has witnessed significant advancements, yielding a variety of detectors~\cite{girshick2015fast,liu2016ssd,ren2017faster,cai2018cascade, he2017mask, lin2017feature, tian2019fcos, carion2020end, zhu2020deformable}. While these detectors have exhibited impressive performance on various public benchmarks~\cite{lin2014microsoft, everingham2010pascal, kuznetsova2020open, shao2019objects365}, their reliance on fully annotated training data poses challenges in detecting objects from unknown classes that are not presented in the training data, hindering their real-world applications.
To remedy this critical limitation, recently, research efforts have been devoted to open-set object detection (OSOD)~\cite{dhamija2020overlooked}, for which the detector, trained on closed-set datasets, is not only tasked with detecting objects from known categories but also objects from unknown categories.

Recent years have witnessed significant advancements in OSOD and its neighboring problem Open World Object Detection (OWOD). While OSOD focuses on detecting both known and unknown objects in a single step, OWOD addresses a similar challenge but in a multi-step context. 
Many methods are proposed based on different detection architectures such as Faster R-CNN~\cite{ren2017faster} and DETR~\cite{carion2020end}.
Existing approaches can be broadly categorized into two streams according to their strategies for handling unlabeled instances.
The first stream~\cite{han2022expanding, zohar2023prob} focuses on learning universal features from only the annotated instances to generalize knowledge from closed-set to open-set classes. 
The second stream~\cite{joseph2021towards, gupta2022ow, ma2023cat, wang2023random, sun2024exploring} proactively leverages unlabeled instances through pseudo-labeling, assigning the ``unknown'' label to high-confidence proposals that do not overlap with ground-truth boxes.  Thanks to the additional proxy supervision from the unlabeled data, the second stream methods often demonstrate stronger performance than the first stream.

Among existing methods, approaches~\cite{han2022expanding} based on Faster R-CNN have surprisingly demonstrated superior performance over other methods, including the methods based on Transformers. Meanwhile, we also find that in OWOD, methods based on Faster R-CNN also stand as the state-of-the-art~\cite{sun2024exploring}. 
Faster R-CNN breaks the object detection problem into two stages: a class-agnostic detection stage, followed by a class-specific detection refinement stage. The class-agnostic detection stage, achieved by the Region Proposal Network (RPN), naturally offers the capabilities to leverage the unlabeled objects present in the training images, through pseudo-labeling. 
However, through careful experimental analysis, we reveal that the unknown recall of the state-of-the-art open-set detector~\cite{han2022expanding} significantly lags behind that of the RPN in the standard Faster-RCNN. 
This gap suggests that the full potential of Faster R-CNN remains untapped. 
Meanwhile, we identified two critical limitations in existing methods that utilize unlabeled instances due to their method design: (1) pure background regions or partial object parts may be incorrectly labeled as whole unknown objects (\eg, the ``grass'' background in Fig.~\ref{fig:intro}(a) and the bear head in Fig.~\ref{fig:intro}(b)), leading to low-quality detections; (2) unknown objects overlapping with ground-truth boxes are mistakenly treated as background during training (\eg, the ``tennis racket'' co-occurring with ``people'' in Fig.~\ref{fig:intro}(c)). These limitations significantly restrict their performance in real-world scenarios where unknown objects frequently co-occur with known objects.


To unleash the full potential of the two-stage design inherited from the classic and strong Faster R-CNN models for OSOD and address the aforementioned two major limitations, we propose a remarkably simple yet effective framework for OSOD, called UADet, short for \textbf{U}ncertainty-\textbf{A}ware Open-Set Object \textbf{Det}ector. 
Our method introduces uncertainty awareness when leveraging the unknown objects in the training images, taking into account both the appearance uncertainty (\emph{i.e.}, the likelihood of an object being in the foreground) and the geometric uncertainty (\emph{i.e.}, the amount of overlapping with known objects). 
By integrating these factors, our method significantly increases the efficacy of leveraging the unannotated unknown objects in the training data. UADet greatly improves the unknown recall and effectively mitigates the issue of overconfident predictions for false positives of unknown objects while effectively utilizing unlabeled unknown instances, leading to a substantial performance improvement. 

We thoroughly evaluate our method on public benchmarks for OSOD, achieving a $1.8\times$ increase in unknown recall while maintaining a higher mean Average Precision (mAP) for known classes than the previous state-of-the-art~\cite{han2022expanding}. Moreover, we also extend the evaluation to the OWOD task, obtaining superior performance over the previous state-of-the-art~\cite{sun2024exploring}, with average improvements of 13.8\% and 6.9\% in unknown recall on M-OWODB and S-OWODB benchmarks respectively. This results in an overall 10.4\% gain. Our UADet establishes the new state-of-the-art on both OSOD and OWOD. 

In summary, we make the following three contributions:
\textit{First}, we find that existing OSOD methods struggle with recalling unknown objects, and we identify the major causes of ineffectiveness: inadequate utilization and improper handling of unlabeled instances.
\textit{Second}, we propose UADet, an uncertainty-aware open-set detection framework that incorporates both appearance and geometric uncertainties to better leverage unlabeled unknown objects during training.
\textit{Third}, extensive experiments on public benchmarks demonstrate that our method achieves state-of-the-art performance on both OSOD and OWOD tasks.

    
    


\section{Related Work}
\label{sec:related}

\noindent {\bf Open-Set Recognition and Detection.}
The Open-Set Recognition (OSR) problem was initially defined as a constrained minimization problem in~\cite{scheirer2012toward} and later extended to multi-class classifiers in subsequent works~\cite{jain2014multi,scheirer2014probability, junior2017nearest, zhang2016sparse}.  The field witnessed a significant advancement with OpenMax~\cite{bendale2016towards}, the first deep learning-based approach that leverages feature space analysis and ensemble risk estimation for unknown detection. Subsequent methods explored diverse strategies: prototype learning~\cite{chen2020learning,chen2021adversarial}, autoencoder-based reconstruction~\cite{yoshihashi2019classification,oza2019c2ae,sun2020conditional}, and generative modeling~\cite{ge2017generative,neal2018open}. Extending OSR to detection, Dhamija~\emph{et al.}~\cite{dhamija2020overlooked} formalized Open-Set Object Detection (OSOD) after observing detectors' tendency to misclassify unknown objects. Their evaluation showed that background-aware detectors~\cite{ren2017faster} outperformed one-vs-rest~\cite{lin2017focal} and objectness-based~\cite{redmon2016you} approaches. Recent OSOD advances focus on uncertainty quantification~\cite{miller2018dropout,miller2019evaluating}, including Dropout sampling~\cite{miller2018dropout,gal2016dropout} and Gaussian Mixture modeling~\cite{miller2021uncertainty}. More recently, OpenDet~\cite{han2022expanding} proposes to identify unknown objects by separating high/low-density regions in the latent space. However, its reliance on closed-set training data and limited utilization of unlabeled unknown instances limits its unknown-aware capabilities.

\noindent {\bf Open-world Object Detection.} 
Joseph et al. \cite{joseph2021towards} propose the Open-world Object Detection (OWOD) task, extending the OSOD task to a dynamic scenario where the model should recognize known and unknown objects while being incrementally trained with new knowledge. Recent research efforts in this area can be broadly categorized into two streams based on whether they utilize unlabeled unknown instances in the training data. The first stream~\cite{ma2023annealing, zohar2023prob} relies solely on annotated instances, attempting to learn universal features from known classes to generalize to unknown classes. For instance, PROB~\cite{zohar2023prob} models the probability of foreground objects using a class-agnostic multivariate Gaussian distribution. Although these methods show promise, their limited use of training data results in suboptimal unknown recall. The second stream~\cite{joseph2021towards, gupta2022ow, wu2022uc, yang2021objects, wang2023random, ma2023cat, zheng2022towards,sun2024exploring} actively leverages unlabeled objects through pseudo-labeling strategies, typically identifying unknown objects as high-objectness proposals without ground-truth overlap. Representative approaches include ORE~\cite{joseph2021towards} utilizing RPN-based objectness scores, OW-DETR~\cite{gupta2022ow} leveraging encoder activation maps, and Randbox~\cite{wang2023random} computing objectness from logit summation. More recently, OrthogonalDet~\cite{sun2024exploring} addresses the correlation between objectness and class information through orthogonality constraints.  However, these methods still face challenges in unknown object recall, stemming from both inadequate utilization and improper handling of unlabeled unknown instances in the training data.

\section{Method}
\label{sec:method}

\noindent{\bf Problem Statement.} Consider an object detection dataset with $N$ images $\{\mathbf{I}_1,...,\mathbf{I}_N\}$ and their corresponding labels $\{\mathbf{Y}_1,...,\mathbf{Y}_N\}$. Each label $\mathbf{Y}_i=\{\mathbf{y}_1,...,\mathbf{y}_M\}$ consists of $M$ object annotations, where $i\in[1,2,..., N]$ and  $\mathbf{y}_m=[l_m,x_m,y_m,w_m,h_m]$ is the object annotation 
containing the class label $l_m$ and the bounding box annotations $x_m,y_m,w_m,h_m$, $m\in[1,2,..., M]$. During training, the model is trained on data containing $K$ known classes $C_\mathcal{K}=\{1, ..., K\}$, represented as $\{(\mathbf{I}_n,\mathbf{Y}_n) | l_k\in C_{\mathcal{K}},\mathbf{y}_k\in\mathbf{Y}_{n}\}_{n=1}^{N^{tr}}$. During testing, the model is evaluated on data that includes known classes $C_{\mathcal{K}}$ and unknown classes $C_{\mathcal{U}}$, denoted as $\{(\mathbf{I}_n,\mathbf{Y}_n) | l_k\in C_{\mathcal{K}}\cup C_{\mathcal{U}},\mathbf{y}_k\in\mathbf{Y}_{n}\}_{n=1}^{N^{te}}$. Here, $N^{tr}$ and $N^{te}$ represent the number of images in the training set and test set, respectively, and $N = N^{tr}+N^{te}$. The goal of OSOD is to localize and recognize objects from both known classes $C_\mathcal{K}$ and unseen classes $C_{\mathcal{U}}$.


\noindent {\bf Preliminary.} In the standard Faster R-CNN framework, the first stage consists of a class-agnostic detector including an image encoder $\Phi_{enc}$ and a RPN $\Phi_{rpn}$. For an image $\mathbf{I}_i$, the RPN generates $N_{i}$ region proposals denoted as $\hat{r}_{i} = \{ \hat{r}_{i,j}\}_{j=1}^{N_{i}}$, along with their respective features $\hat{f}_{i,j}$ and objectness scores $\hat{o}_{i,j}$, where $i \in \{1, 2, ..., N\}$, $j \in \{1, 2, ..., N_{i}\}$. The RPN process can be expressed as:
\begin{equation*}
{\{\hat{r}_{i,j}, \hat{f}_{i,j}, \hat{o}_{i,j}}\}_{j=1}^{N_{i}} = {\Phi_{rpn}} \circ \Phi_{enc}(\mathbf{I}_i)
\end{equation*}

The second stage consists of a RoI head $\Phi_{roi}$, a classification head $\Phi_{cls}$, and a regression head $\Phi_{box}$. The output of these heads for proposal $r_{i, j}$ can be formulated as:
\begin{align*}
 o_{i, j} &= \Phi_{cls} \circ \Phi_{roi}(\{\hat{r}_{i,j}, \Phi_{enc}(\mathbf{I}_i)) \\
 b_{i, j} &= \Phi_{box} \circ \Phi_{roi}(\{\hat{r}_{i,j}, \Phi_{enc}(\mathbf{I}_i))
\end{align*}

In both stages, proposals are matched with GT boxes based on their Intersection over Union (IoU) score, denoted as $u_{i, j}$. Proposals with IoU scores above a threshold (\eg, 0.5) are considered positive samples for the corresponding object category, while those below are treated as negative samples. The predicted confidence score $p_{i,j}^{c}$ for class $c$ and the final losses can be expressed as:
\begin{equation} 
    p_{i,j}^{c} = \text{softmax}(o_{i,j}^{c}) = \frac{\exp(o_{i,j}^{c})}{\sum_{k \in C}\exp(o_{i,j}^{k})},
\end{equation}
where $C = C_{\mathcal{K}} \cup C_{bg}$ and $C_{bg}=\{K+1\}$. The classification and box regression losses are:
\begin{equation}
L_{cls}(p_{i,j}, y_{i,j}) = -\sum_{c \in C} y_{i,j}^{c} \log(p_{i,j}^{c}),
\end{equation}
\begin{equation}
L_{reg}(b_{i,j}, t_{i,j}) = \text{Smooth-L1}(b_{i,j}, t_{i,j}),
\end{equation}
where $y_{i,j}$ and $t_{i,j}$ represent the GT labels and box annotations respectively, and $y_{i,j} \in \{0, 1\}$.

\subsection{Motivation: Recall Analysis}
\begin{figure}[t]
    \centering
    \includegraphics[width=0.3\textwidth]{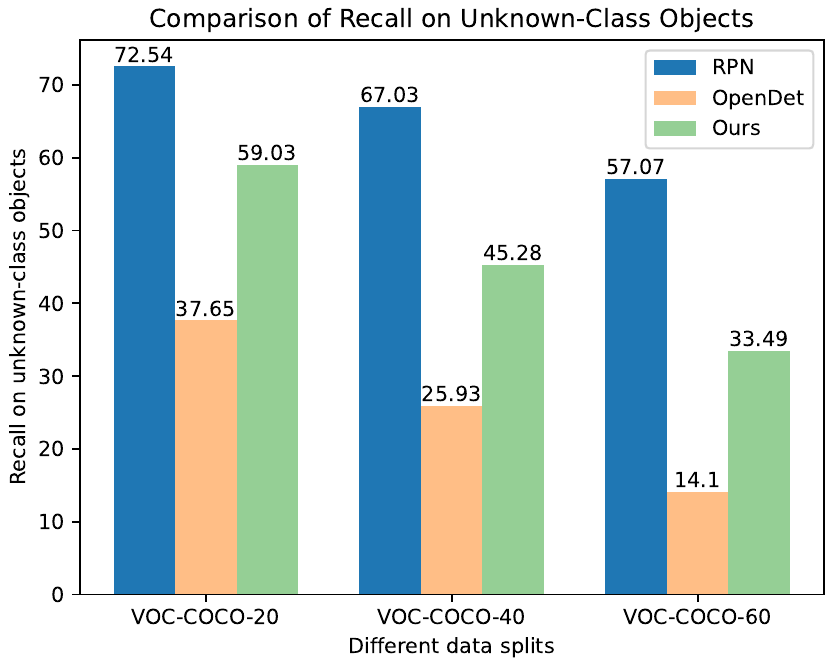}
    \caption{\textbf{The comparison of unknown recall among RPN, OpenDet, and our proposed method.} The evaluation was performed on three distinct data splits, \emph{i.e.}, VOC-COCO-\{20,40,60\} introduced in~\cite{han2022expanding}. Notably, our proposed method exhibited a significant improvement in terms of unknown recall when compared to the SOTA OSOD framework, OpenDet.}
    \label{fig:motivation}
\end{figure}

Faster R-CNN has gained widespread popularity as a versatile detector across various tasks~\cite{xu2021end,tan2020equalization,wang2020frustratingly, li2022cross}.
Notably, the SOTA detector for OSOD, OpenDet, also builds upon the Faster R-CNN framework. Additionally, the Faster R-CNN-based approaches have also demonstrated superior performance in OWOD, obtaining the state-of-the-art results~\cite{sun2024exploring}.
Faster R-CNN consists of two stages: a class-agnostic RPN that generates multiple proposals with their corresponding probabilities of being foreground objects, and a class-specific detector responsible for category prediction and bounding box regression using the RPN's proposals.
Given the RPN's capability to generate a vast number of proposals, which are likely to cover all potential foreground objects, it is anticipated that the Faster R-CNN-like framework would achieve high recall for unknown classes in the OSOD task. However, surprisingly, when comparing the unknown recall of the RPN with OpenDet, the current SOTA method, OpenDet exhibited significantly lower recall rates, as shown in Fig.~\ref{fig:motivation}. 
This discrepancy highlights the untapped potential of the Faster R-CNN-like architecture for OSOD. 
Therefore, we aim to bridge this gap. 



\subsection{UADet: Uncertainty-Aware Open-Set Detector}
We introduce a simple and effective uncertainty-aware framework, \emph{i.e.}, UADet, which adapts the Faster R-CNN model for OSOD. 
UADet incorporates novel uncertainty-guided supervision that takes into account both appearance uncertainty and geometry uncertainty. By harnessing the advantages of these two uncertainty measures, our approach effectively leverages the capabilities of the RPN, leading to promising open-set performance on both OSOD and OWOD benchmarks.

Previous methods~\cite{joseph2021towards, wu2022uc} attempt to leverage unlabeled instances by identifying high-confidence proposals that do not overlap with GT boxes as unknown objects. These proposals are then utilized to provide additional supervision for unknown object detection through the classification head. Let us denote $q_{i,j}$ as the label assigned for each proposal, and the classification loss can be formulated as:
\begin{equation} \label{eq:ce_loss_osod}
L_{cls}(p_{i,j}, q_{i,j}) = -\sum_{c \in C^{'}} q_{i,j}^{c} \log(p_{i,j}^{c}),
\end{equation}
where $C^{'} = C_{\mathcal{K}} \cup C_{\mathcal{U}} \cup C_{bg}$ and $q_{i,j}\ \in \{0, 1\}$. Given the unconstrained nature of unknown classes in OSOD, we unify them into a single category, \emph{i.e.}, $C_{\mathcal{U}}=\{K+1\}$ and $C_{bg}=\{K+2\}$.

While existing methods that leverage unlabeled instances through pseudo-labels often demonstrate stronger performance than those that do not, they have several critical limitations (Fig.~\ref{fig:intro}). 
Firstly, the selected high-confidence samples might only encompass the pure background, leading to a potential blurring of the boundary between unknown objects and pure background. Secondly, these samples may only cover partial parts of unknown objects, which can result in numerous low-quality predictions when assigned hard labels. Finally, these methods focus solely on proposals with no overlapping with the GT boxes ($u_{i,j}=0$), neglecting those with partial overlap ($u_{i,j} \in [0, 0.5]$). However, these partially overlapping proposals could contain unknown objects and provide valuable supervision for unknown object detection. These limitations significantly restrict their performance in real-world scenarios where unknown objects \textit{frequently co-occur} with known objects.

\label{subsec:The_proposed_method}
\begin{figure}[t]
    \centering
    \includegraphics[width=0.7\linewidth]{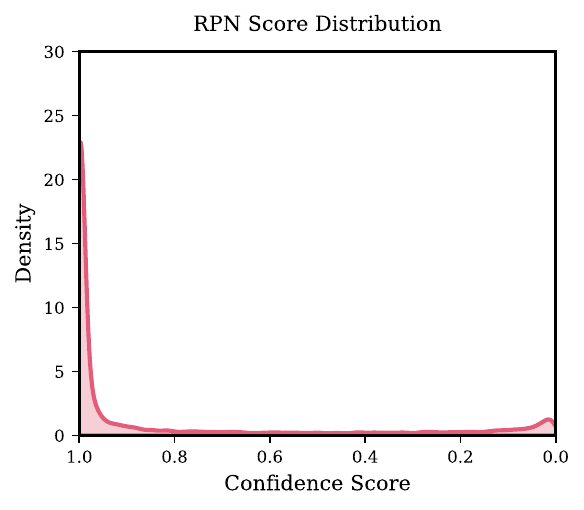}
    \caption{\textbf{The unknown score distribution histograms of the RPN in a closed-set trained Faster R-CNN model.} The statistics is done on VOC-COCO-20~\cite{han2022expanding}.}
    \label{fig:RPN_conf}
\end{figure}

\subsubsection{Uncertainty-Awareness Modeling} 

To address the above issues, we propose a simple and effective uncertainty-aware method for OSOD that meets the following criteria: (i) Pure background should not be mistakenly considered unknown objects. (i) Proposals covering partial parts of unknown objects should receive soft supervision instead of hard ones. (iii) Negative proposals should be fully utilized to leverage potential unknown objects in unlabeled data. These considerations have led us to develop our uncertainty-based detection framework. 

Notably, experiments demonstrate that the RPN score can effectively reflect the confidence of the foreground object distribution, as depicted in Fig~\ref{fig:RPN_conf}. A histogram analysis of the RPN-recalled unknown objects reveals that a majority of these objects have high objectness scores ([0.8, 1]), with proportions of 78.5\% on VOC-COCO-20, which are sufficient to bring significant gains. This clearly indicates that the RPN's objectness score can serve as a reliable measure for uncertainty estimation.

Inspired by this plausible characteristic of the RPN's objectness score, we treat it as an indicator of \textit{appearance uncertainty} and assume that the probability of a negative proposal containing an unknown foreground object is proportional to $\hat{o}_{i, j}$. Consequently, we can express the assigned label $s_{i,j}^{K+1}$ for the negative proposal $r_{i,j}$ as follows: 
\begin{equation} \label{eq:ce_loss}
s_{i,j}^{K+1} \propto \hat{o}_{i, j}.
\end{equation}

\begin{figure}[t]
    \centering
    \includegraphics[width=1.0\linewidth]{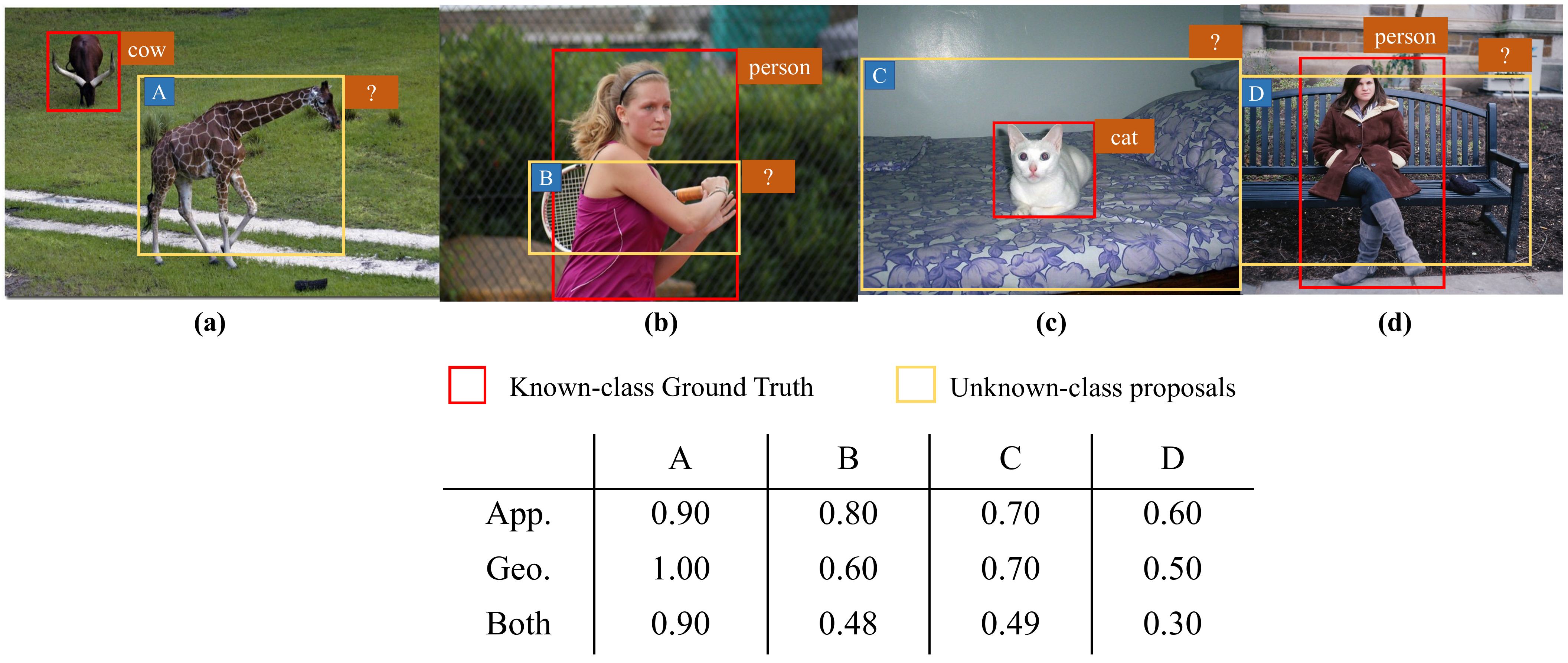}
    \caption{\textbf{An illustration of the necessity to introduce geometric uncertainty.} We present four negative proposals generated by the RPN. \text{App.} and \text{Geo.} represent the appearance uncertainty and the geometry uncertainty. For simplicity, we set App.= $\hat{o}_{i,j}$, Geo.= $1 - u_{i,j}$,  Both = $\text{App.} \times \text{Geo}$. Among these proposals, B, C, and D exhibit ambiguity. By incorporating geometric uncertainty, we mitigate the reliance solely on appearance uncertainty during the label assignment process, thereby mitigating the potential decline in performance for known classes. }
    \label{fig:geo_uncer}
    \vspace{-3mm}
\end{figure}

However, it should be noted that some negative proposals may contain partial parts of known objects, and assigning their label solely based on their confidence scores is likely to degrade the performance of known classes. To illustrate this point, we present four examples in Fig.~\ref{fig:geo_uncer}. For proposal A, which has no overlap with GT boxes, the uncertainty estimation based on appearance will not cause any interference to the prediction of the known class (\emph{i.e.}, the cow in Fig.~\ref{fig:geo_uncer}(a)). However, proposals B, C, and D encompass regions that include both known and unknown objects, resulting in overlapping regions on their corresponding feature maps. This overlap can introduce ambiguity in the model's predictions, as the overlapping features are forced to contribute to the prediction of both known and unknown classes simultaneously. Therefore, relying solely on appearance uncertainty scores may lead to some known objects being misclassified as unknown objects.

To address this issue, we propose to utilize the IoU of the proposal and its corresponding GT box as the measure of \textit{geometry uncertainty}. We set $s_{i,j}^{K+1}$ to be inversely correlated to the IoU score $u_{i,j}$, which can be expressed as:
\begin{equation} \label{eq:geo_un}
s_{i,j}^{K+1} \propto 1 - u_{i,j}.
\end{equation}
From Eq.\ref{eq:geo_un}, we can observe that the larger the IoU between a negative proposal and the GT of the known class, the lower its geometric uncertainty score, which in turn reduces the interference to the known class predictions caused by supervision signal generated solely based on the appearance uncertainty score. Specifically, as shown in Fig.\ref{fig:geo_uncer}, after incorporating geometric uncertainty, the model's confidence in predicting proposal B, C, and D as unknown classes was correspondingly reduced. Moreover, our experiments in Sec.\ref{subsec:ablation} also validate that the proposed geometric uncertainty score can effectively mitigate the potential decline in known class performance.



Based on the above analysis, we propose to incorporate both the appearance uncertainty measure $o_{i, j}$ and the geometry uncertainty measure $u_{i,j}$ into our label assignment strategy for negative proposals. We can then express the soft label $s_{i,j}$ for each negative proposal $r_{i,j}$ as:
\begin{equation}
    s_{i,j}^{c} = \begin{cases}
    0, & \text{if } c \in \{1, 2, ... , K\} \\
    \hat{o}_{i, j}(1 - u_{i,j}), & \text{if } c = K + 1\\
    1 - \hat{o}_{i, j}(1 - u_{i,j}), & \text{if }  c = K + 2 
    \end{cases}
\end{equation}


Denote $q_{i,j}$ as the assigned label for each proposal $r_{i,j}$ with our proposed uncertainty-awareness modeling. We can then formulate it as:
\begin{equation} 
    q_{i,j}^{c} = \begin{cases}
    y_{i,j}^{c}, & \text{if } r_{i,j} \text{ is a positive proposal} \\
    s_{i,j}^{c}, & \text{if } r_{i,j} \text{ is a negative proposal}
    \end{cases},
\end{equation}
which can be substituted into the Eq.~\ref{eq:ce_loss_osod} to obtain the final classification loss.

\subsubsection{Overall Optimization}
\label{subsec:overall_optim}
Guided by the uncertainty score above, our model can be trained by the following multi-task loss:
\begin{equation}
    \mathcal{L} = \mathcal{L}_{rpn} + \mathcal{L}_{reg} + \mathcal{L}_{cls},
\end{equation}
where $\mathcal{L}_{rpn}$ denotes the total loss of RPN including the regression and classification loss, $\mathcal{L}_{reg}$ is smooth L1 loss for box regression, and $\mathcal{L}_{cls}$ is classification loss for box classification. Note that our uncertainty awareness is only used during training. After training, the model can be directly applied for OSOD in a feed-forward pass, similar to Faster-RCNN, highlighting the great simplicity of our method.

\subsubsection{Extension to OWOD}

Thanks to the simplicity of our method, it can be easily extended to OWOD by incorporating exemplar replay-based fine-tuning~\cite{joseph2021towards, gupta2022ow, yang2021objects}. This approach helps mitigate catastrophic forgetting of previously learned classes. Specifically, after each incremental step in an episode, the model is fine-tuned using a balanced set of exemplars stored for each known class.

\section{Experiment}
\label{sec:exp}

\begin{table*}[t]
    \caption{\textbf{Comparisons with other Methods on VOC and VOC-COCO-T$_1$.} $^*$ denotes that the experimental results are obtained from the log file provided by OpenDet's official code repository, as the original OpenDet paper does not report the unknown recall. $^\dag$ indicates that this is the result of our reproduction. The second highest ranked result is underlined.}
    \centering
\small
\resizebox{\textwidth}{!}{%
\begin{tabular}{l|c|ccccc|ccccc|ccccc@{}}
\midrule
\multirow{2}{*}{Method} & VOC & \multicolumn{5}{c|}{VOC-COCO-20} & \multicolumn{5}{c|}{VOC-COCO-40} & \multicolumn{5}{c}{VOC-COCO-60} \\ \cmidrule(lr){2-2} \cmidrule(lr){3-7}  \cmidrule(lr){8-12} \cmidrule(lr){13-17}
                & mAP ($\uparrow$) & \cellcolor[HTML]{FFFFED}{U-Recall ($\uparrow$)} & \cellcolor[HTML]{FFFFED}mAP ($\uparrow$) & WI ($\downarrow$)   & AOSE ($\downarrow$) & U-AP ($\uparrow$)  & \cellcolor[HTML]{FFFFED}{U-Recall ($\uparrow$)} &  \cellcolor[HTML]{FFFFED}mAP ($\uparrow$) & WI ($\downarrow$)   & AOSE ($\downarrow$)  & U-AP ($\uparrow$)   & \cellcolor[HTML]{FFFFED}{U-Recall ($\uparrow$)}  & \cellcolor[HTML]{FFFFED}mAP ($\uparrow$)  & WI ($\downarrow$) & AOSE ($\downarrow$) & U-AP ($\uparrow$) \\ \midrule
    \multicolumn{17}{l}{\textit{Fully-supervised ImageNet-pretrained ResNet50 as Backbone:}} \\ 
    \hline
    FR-CNN$^*$~\cite{ren2017faster}    & 80.06 & \cellcolor[HTML]{FFFFED}0 & \cellcolor[HTML]{FFFFED}58.36 & 19.50 & 16518 & 0 & \cellcolor[HTML]{FFFFED}0  & \cellcolor[HTML]{FFFFED}55.34 & 23.78  & 25339 & 0 & \cellcolor[HTML]{FFFFED}0 & \cellcolor[HTML]{FFFFED}55.94  & 18.91 & 27255 & 0 \\ 
    PROSER$^*$~\cite{zhou2021learning} & 79.42 & \cellcolor[HTML]{FFFFED}37.34 & \cellcolor[HTML]{FFFFED}56.72 & 20.44 & 14266 & \bf{16.99}  & \cellcolor[HTML]{FFFFED}25.10 & \cellcolor[HTML]{FFFFED}53.48  & 25.28 & 21992  & \bf{11.76} & \cellcolor[HTML]{FFFFED}13.02 & \cellcolor[HTML]{FFFFED}54.57  & 20.18 & 23217 & \underline{4.56}  \\
    ORE~\cite{joseph2021towards}   & 79.80 & \cellcolor[HTML]{FFFFED}- & \cellcolor[HTML]{FFFFED}58.25 & 18.18 & 12811 & 2.60 & \cellcolor[HTML]{FFFFED}- & \cellcolor[HTML]{FFFFED}55.30 & 22.40 & 19752  & 1.70 & \cellcolor[HTML]{FFFFED}- & \cellcolor[HTML]{FFFFED}55.47 & 18.35 & 21415 & 0.53    \\ 
    DS$^*$~\cite{miller2018dropout}    & 79.70 & \cellcolor[HTML]{FFFFED}19.80 & \cellcolor[HTML]{FFFFED}58.46 & 16.76 & 13062 & 8.75  & \cellcolor[HTML]{FFFFED}12.93 & \cellcolor[HTML]{FFFFED}55.28 & 20.34 & 19891 & 5.94 & \cellcolor[HTML]{FFFFED}5.57  & \cellcolor[HTML]{FFFFED}56.19 & 16.77 & 22406& 2.09 \\ 
    OpenDet$^*$~\cite{han2022expanding} & \underline{80.02} & \cellcolor[HTML]{FFFFED}\underline{37.65} & \cellcolor[HTML]{FFFFED}\underline{58.64} & \bf{12.50} & \underline{10758} & 14.38 & \cellcolor[HTML]{FFFFED}\underline{25.93} & \cellcolor[HTML]{FFFFED}\underline{55.60} & \bf{15.38} & \underline{16061} & 10.49 & \cellcolor[HTML]{FFFFED}\underline{14.10} & \cellcolor[HTML]{FFFFED}\underline{56.12} & \bf{12.76} & \underline{18548} & 4.37 \\
    \rowcolor{lightyellow}
    \bf{UADet} & \bf{80.13} & \cellcolor[HTML]{FFFFED}\bf{59.03} & \cellcolor[HTML]{FFFFED}\bf{59.12} & \underline{13.19} & \bf{10186} & \underline{15.09} & \cellcolor[HTML]{FFFFED}\bf{45.28} & \cellcolor[HTML]{FFFFED}\bf{56.00} & \underline{16.17} &  \bf{14956} & \underline{11.51} & \cellcolor[HTML]{FFFFED}\bf{33.49} & \cellcolor[HTML]{FFFFED}\bf{56.73}  & \underline{12.77} & \bf{16550} &  \bf{5.97} \\
    \midrule
    \multicolumn{17}{l}{\textit{Self-supervised DINO-pretrained ResNet50 as Backbone:}} \\
    \hline
    OpenDet$^\dag$~\cite{han2022expanding} & \underline{82.12} & \cellcolor[HTML]{FFFFED}40.76  & \cellcolor[HTML]{FFFFED}\underline{62.09}  & \bf{12.03} &  12486 & \underline{17.05} & \cellcolor[HTML]{FFFFED}27.26 & \cellcolor[HTML]{FFFFED}\underline{59.32} & \bf{14.47} & 18219 & 12.02 & \cellcolor[HTML]{FFFFED}14.90& \cellcolor[HTML]{FFFFED}\underline{60.18}  &\bf{11.52} &  21969 & 4.81\\ 
    
    PROB$^\dag$~\cite{zohar2023prob} & 81.77 & \cellcolor[HTML]{FFFFED}\underline{45.61}  & \cellcolor[HTML]{FFFFED}59.91 & 18.38 & \bf{10408} & 4.92 & \cellcolor[HTML]{FFFFED}\underline{32.41} & \cellcolor[HTML]{FFFFED}55.71 & 22.02 & \bf{14935} & \underline{10.60} & \cellcolor[HTML]{FFFFED}\underline{19.61} & \cellcolor[HTML]{FFFFED}56.55  & 15.98 &  \bf{14333} & \underline{5.14} \\

    \rowcolor{lightyellow}
    \bf{UADet} & \textbf{82.25} & \cellcolor[HTML]{FFFFED}\bf{67.64}  & \cellcolor[HTML]{FFFFED}\bf{62.91} & \underline{12.96} & \underline{12351} & \bf{17.55} & \cellcolor[HTML]{FFFFED}\bf{55.66} & \cellcolor[HTML]{FFFFED}\bf{60.16} & \underline{15.32} & \underline{18178} & \bf{13.42} & \cellcolor[HTML]{FFFFED}\bf{44.21} & \cellcolor[HTML]{FFFFED}\bf{60.98}  & \underline{12.03} & \underline{21531} & \bf{7.40} \\
    \midrule
\end{tabular}%
}
    \label{tab:voc_coco_a}
\end{table*}

\begin{table*}[t]
    \caption{\textbf{Comparisons with other Methods on VOC and VOC-COCO-T$_2$.} $^*$ denotes that the experimental results are obtained from the log file provided by OpenDet's official code repository, as the original OpenDet paper does not report the unknown recall. $^\dag$ indicates that this is the result of our reproduction. The second highest ranked result is underlined.}
    \centering
\small
\resizebox{\textwidth}{!}{%
\begin{tabular}{l|ccccc|ccccc|ccccc}
\midrule
\multirow{2}{*}{Method} & \multicolumn{5}{c|}{VOC-COCO-2500} & \multicolumn{5}{c|}{VOC-COCO-5000} & \multicolumn{5}{c}{VOC-COCO-20000} \\ \cmidrule(lr){2-6}  \cmidrule(lr){7-11} \cmidrule(lr){12-16}
                 & \cellcolor[HTML]{FFFFED}U-Recall ($\uparrow$) & \cellcolor[HTML]{FFFFED}mAP ($\uparrow$)  & WI ($\downarrow$)   & AOSE ($\downarrow$) & U-AP ($\uparrow$)  &  \cellcolor[HTML]{FFFFED}U-Recall ($\uparrow$) & \cellcolor[HTML]{FFFFED}mAP ($\uparrow$) & WI ($\downarrow$)   & AOSE ($\downarrow$) & U-AP ($\uparrow$) & \cellcolor[HTML]{FFFFED}U-Recall ($\uparrow$)  & \cellcolor[HTML]{FFFFED}mAP ($\uparrow$) &
                WI ($\downarrow$)   & AOSE ($\downarrow$)  & U-AP ($\uparrow$) \\ \midrule
\multicolumn{7}{l}{\textit{Fully-supervised ImageNet-pretrained ResNet50 as Backbone:}} \\ 
\hline
FR-CNN$^*$~\cite{ren2017faster}    & \cellcolor[HTML]{FFFFED}0  & \cellcolor[HTML]{FFFFED}77.83 & 9.12  & 6424 & 0 & \cellcolor[HTML]{FFFFED}0 & \cellcolor[HTML]{FFFFED}74.61 & 16.63 & 13499  & 0 & \cellcolor[HTML]{FFFFED}0 & \cellcolor[HTML]{FFFFED}64.50 & 33.90 & 52624 & 0    \\ 
PROSER$^*$~\cite{zhou2021learning} & \cellcolor[HTML]{FFFFED}28.57 & \cellcolor[HTML]{FFFFED}76.76  & 9.68 & 5554 & \bf{10.98} & \cellcolor[HTML]{FFFFED}13.27 & \cellcolor[HTML]{FFFFED}72.75  & 28.38 & 11669  & \bf{17.48} & \cellcolor[HTML]{FFFFED} 28.55  & \cellcolor[HTML]{FFFFED}61.52 &35.64 & 45507 & 16.74   \\
ORE~\cite{joseph2021towards}   & \cellcolor[HTML]{FFFFED}- & \cellcolor[HTML]{FFFFED}77.84 & 8.39  & 4945  & 1.75 & \cellcolor[HTML]{FFFFED}-  & \cellcolor[HTML]{FFFFED}74.34 & 15.36 & 10568  & 1.81 & \cellcolor[HTML]{FFFFED}- & \cellcolor[HTML]{FFFFED}64.59 & 32.40 & 40865 & 2.14   \\ 
DS$^*$~\cite{miller2018dropout}  & \cellcolor[HTML]{FFFFED}16.47 & \cellcolor[HTML]{FFFFED}77.69  & 8.28  & 4917 & 4.76 & \cellcolor[HTML]{FFFFED}16.49 & \cellcolor[HTML]{FFFFED}74.29 & 15.16 & 10226 & 6.73 & \cellcolor[HTML]{FFFFED}16.73 & \cellcolor[HTML]{FFFFED}64.02 & 31.50 & 39870 & 9.90   \\ 
OpenDet$^*$~\cite{han2022expanding} & \cellcolor[HTML]{FFFFED}\underline{30.87} & \cellcolor[HTML]{FFFFED}\underline{78.65} & \underline{6.14} & \underline{3781} & 8.69 & \cellcolor[HTML]{FFFFED}\underline{30.46} & \cellcolor[HTML]{FFFFED}\underline{75.38} & \underline{11.07} & \underline{7819} & 12.21 & \cellcolor[HTML]{FFFFED}\underline{30.90} & \cellcolor[HTML]{FFFFED}\bf{65.54} & \bf{24.43} & \underline{30500} & \underline{17.02} \\

 \rowcolor{lightyellow}
\textbf{UADet} & \cellcolor[HTML]{FFFFED}\bf{51.92} & \cellcolor[HTML]{FFFFED}\bf{78.66} & \bf{5.93} &  \bf{3476} & \underline{10.02} &  \cellcolor[HTML]{FFFFED}\bf{51.12} & \cellcolor[HTML]{FFFFED}\bf{75.40} & \bf{10.82} & \bf{7308} & \underline{13.18} & \cellcolor[HTML]{FFFFED}\bf{51.58} & \cellcolor[HTML]{FFFFED}\underline{65.43} & \underline{25.10} & \bf{28536} & \bf{17.87} \\
\midrule
\multicolumn{7}{l}{\textit{Self-supervised DINO-pretrained ResNet50 as Backbone:}} \\
\hline
OpenDet$^\dag$~\cite{han2022expanding} & \cellcolor[HTML]{FFFFED}32.09 & \cellcolor[HTML]{FFFFED}\underline{81.17} & \underline{5.44} & 4285 & \underline{9.47} &  \cellcolor[HTML]{FFFFED}31.87 & \cellcolor[HTML]{FFFFED}\underline{78.13} & \underline{9.87} & 8811 & \underline{12.90} & \cellcolor[HTML]{FFFFED}32.06 & \cellcolor[HTML]{FFFFED}\underline{68.84} & \underline{24.88} & 34424 & \underline{17.95} \\

PROB$^\dag$~\cite{zohar2023prob} & \cellcolor[HTML]{FFFFED}\underline{37.42} & \cellcolor[HTML]{FFFFED}77.64 & 6.35 &  \bf{3273} & 2.02 &  \cellcolor[HTML]{FFFFED}\underline{36.38} & \cellcolor[HTML]{FFFFED}74.51 & 12.46 & \bf{6788} & 10.67 & \cellcolor[HTML]{FFFFED}\underline{37.10} & \cellcolor[HTML]{FFFFED}65.79 & 30.59 & \bf{26744} & 12.04 \\

 \rowcolor{lightyellow}
\bf{UADet}  & \cellcolor[HTML]{FFFFED}\bf{63.07} & \cellcolor[HTML]{FFFFED}\bf{81.60} & \bf{5.01} & \underline{4230} & \bf{11.38} &  \cellcolor[HTML]{FFFFED}\bf{61.57} &  \cellcolor[HTML]{FFFFED}\bf{78.36}  & \bf{9.19}  & \underline{8766} & \bf{15.00} & \cellcolor[HTML]{FFFFED}\bf{62.73} &  \cellcolor[HTML]{FFFFED}\bf{68.90} & \bf{22.72} & \underline{34234} & \bf{20.26} \\
\midrule
\end{tabular}%
}
    \label{tab:voc_coco_b}
\end{table*}

\begin{table*}[htbp] 
    \caption{\textbf{State-of-the-art comparison for OWOD on M-OWODB (top) and S-OWODB (bottom).} The comparison is shown in terms of unknown class recall (U-Recall) and known class mAP@0.5 (for previously, currently, and all known objects). Underlined values represent the second best results.}
    \centering
\setlength{\tabcolsep}{3pt}
\adjustbox{width=0.85\textwidth}{
\begin{tabular}{@{}l|cc|cccc|cccc|ccc@{}}
\toprule
 \textbf{Task IDs} ($\rightarrow$)& \multicolumn{2}{c|}{\textbf{Task 1}} & \multicolumn{4}{c|}{\textbf{Task 2}} & \multicolumn{4}{c|}{\textbf{Task 3}} & \multicolumn{3}{c}{\textbf{Task 4}} \\ \midrule
 
& \cellcolor[HTML]{FFFFED}{U-Recall} & \multicolumn{1}{c|}{\cellcolor[HTML]{EDF6FF}{mAP ($\uparrow$)}} & \cellcolor[HTML]{FFFFED}{U-Recall} & \multicolumn{3}{c|}{\cellcolor[HTML]{EDF6FF}{mAP ($\uparrow$)}} & \cellcolor[HTML]{FFFFED}{U-Recall} & \multicolumn{3}{c|}{\cellcolor[HTML]{EDF6FF}{mAP ($\uparrow$)}} & \multicolumn{3}{c}{\cellcolor[HTML]{EDF6FF}{mAP ($\uparrow$)}}  \\

 & \cellcolor[HTML]{FFFFED}($\uparrow$) & \begin{tabular}[c]{@{}c}Current \\ known\end{tabular} & \cellcolor[HTML]{FFFFED}($\uparrow$) & \begin{tabular}[c]{@{}c@{}}Previously\\  known\end{tabular} & \begin{tabular}[c]{@{}c@{}}Current \\ known\end{tabular} & \cellcolor[HTML]{FFFFED}Both & \cellcolor[HTML]{FFFFED}($\uparrow$) & \begin{tabular}[c]{@{}c@{}}Previously \\ known\end{tabular} & \begin{tabular}[c]{@{}c@{}}Current \\ known\end{tabular} & \cellcolor[HTML]{FFFFED}Both & \begin{tabular}[c]{@{}c@{}}Previously \\ known\end{tabular} & \begin{tabular}[c]{@{}c@{}}Current \\ known\end{tabular} & \cellcolor[HTML]{FFFFED}Both \\ \midrule

ORE*~\cite{joseph2021towards} & \cellcolor[HTML]{FFFFED} 4.9  & 56.0 & \cellcolor[HTML]{FFFFED}2.9 & 52.7 & 26.0 & \cellcolor[HTML]{FFFFED}39.4  & \cellcolor[HTML]{FFFFED}3.9 & 38.2 & 12.7 & \cellcolor[HTML]{FFFFED}29.7 & 29.6 & 12.4 & \cellcolor[HTML]{FFFFED}25.3 \\ 

UC-OWOD~\cite{wu2022uc} & \cellcolor[HTML]{FFFFED} 2.4  & 50.7 & \cellcolor[HTML]{FFFFED} 3.4 & 33.1 & 30.5 & \cellcolor[HTML]{FFFFED}31.8  & \cellcolor[HTML]{FFFFED} 8.7 & 28.8 & 16.3 & \cellcolor[HTML]{FFFFED}24.6 & 25.6 & 15.9 & \cellcolor[HTML]{FFFFED}23.2 \\ 

OCPL~\cite{yu2023open} & \cellcolor[HTML]{FFFFED} 8.26  & 56.6 & \cellcolor[HTML]{FFFFED} 7.65 & 50.6 & 27.5 & \cellcolor[HTML]{FFFFED}39.1  & \cellcolor[HTML]{FFFFED} 11.9 & 38.7 & 14.7 & \cellcolor[HTML]{FFFFED}30.7 & 30.7 & 14.4 & \cellcolor[HTML]{FFFFED}26.7 \\ 

2B-OCD~\cite{wu2022two} & \cellcolor[HTML]{FFFFED} 12.1  & 56.4 & \cellcolor[HTML]{FFFFED} 9.4 & 51.6 & 25.3 & \cellcolor[HTML]{FFFFED}38.5  & \cellcolor[HTML]{FFFFED} 11.6 & 37.2 & 13.2 & \cellcolor[HTML]{FFFFED}29.2 & 30.0 & 13.3 & \cellcolor[HTML]{FFFFED}25.8 \\ 

OW-DETR~\cite{gupta2022ow} & \cellcolor[HTML]{FFFFED}7.5  & 59.2 & \cellcolor[HTML]{FFFFED}6.2 & 53.6 & 33.5 & \cellcolor[HTML]{FFFFED}42.9 & \cellcolor[HTML]{FFFFED}5.7 & 38.3 & 15.8 & \cellcolor[HTML]{FFFFED}30.8 & 31.4 & 17.1 & \cellcolor[HTML]{FFFFED}27.8 \\

ALT~\cite{ma2023annealing} & \cellcolor[HTML]{FFFFED}13.6 & 59.3 & \cellcolor[HTML]{FFFFED}10.0 & 53.2 & 34.0 & \cellcolor[HTML]{FFFFED}45.6 & \cellcolor[HTML]{FFFFED}14.3 & 42.6 & 26.7 & \cellcolor[HTML]{FFFFED}38.0 & 33.5 & 21.8 & \cellcolor[HTML]{FFFFED}30.6 \\

PROB~\cite{zohar2023prob} & \cellcolor[HTML]{FFFFED}19.4 & 59.5 & \cellcolor[HTML]{FFFFED}17.4 & \underline{55.7} & 32.2 & \cellcolor[HTML]{FFFFED}44.0 & \cellcolor[HTML]{FFFFED}19.6 & 43.0 & 22.2 & \cellcolor[HTML]{FFFFED}36.0 & 35.7 & 18.9 & \cellcolor[HTML]{FFFFED}31.5 \\

CAT~\cite{ma2023cat} & \cellcolor[HTML]{FFFFED}23.7 & 60.0 & \cellcolor[HTML]{FFFFED}19.1 & 55.5 & 32.7 & \cellcolor[HTML]{FFFFED}44.1 & \cellcolor[HTML]{FFFFED}24.4 & 42.8 & 18.7 & \cellcolor[HTML]{FFFFED}34.8 & 34.4 & 16.6 & \cellcolor[HTML]{FFFFED}29.9 \\

RandBox~\cite{wang2023random} & \cellcolor[HTML]{FFFFED}10.6 & \underline{61.8} & \cellcolor[HTML]{FFFFED}6.3 & - &  & \cellcolor[HTML]{FFFFED}45.3 & \cellcolor[HTML]{FFFFED}7.8 & - & - & \cellcolor[HTML]{FFFFED}39.4 & - & - & \cellcolor[HTML]{FFFFED}35.4 \\

OrthorganalDet~\cite{sun2024exploring} & \cellcolor[HTML]{FFFFED}\underline{24.6} & 61.3 & \cellcolor[HTML]{FFFFED}\underline{26.3} & 55.5 & \underline{38.5} & \cellcolor[HTML]{FFFFED}\underline{47.0} & \cellcolor[HTML]{FFFFED}\underline{29.1} & \underline{46.7} & \textbf{30.6} & \cellcolor[HTML]{FFFFED}\underline{41.3} & \textbf{42.4} & \underline{24.3} & \cellcolor[HTML]{FFFFED}\textbf{37.9} \\

\rowcolor{lightyellow}
\textbf{UADet} & \cellcolor[HTML]{FFFFED}\textbf{38.9} & \textbf{62.3} & \cellcolor[HTML]{FFFFED}\textbf{39.7} & \textbf{58.6} & \textbf{39.1} & \cellcolor[HTML]{FFFFED}\textbf{48.9} & \cellcolor[HTML]{FFFFED}\textbf{42.9} & \textbf{47.0} & \underline{30.1} & \cellcolor[HTML]{FFFFED}\textbf{41.4} & \underline{42.0} & \textbf{24.7} & \cellcolor[HTML]{FFFFED}\textbf{37.9}\\

\hline
\midrule
ORE*~\cite{joseph2021towards} & \cellcolor[HTML]{FFFFED}1.5  & 61.4 & \cellcolor[HTML]{FFFFED}3.9 & 56.5 & 26.1 & \cellcolor[HTML]{FFFFED}40.6  & \cellcolor[HTML]{FFFFED}3.6 & 38.7 & 23.7 & \cellcolor[HTML]{FFFFED}33.7 & 33.6 & 26.3 & \cellcolor[HTML]{FFFFED}31.8 \\ 
OW-DETR~\cite{gupta2022ow} & \cellcolor[HTML]{FFFFED}5.7  & 71.5 & \cellcolor[HTML]{FFFFED}6.2 & 62.8 & 27.5 & \cellcolor[HTML]{FFFFED}43.8 & \cellcolor[HTML]{FFFFED}6.9 & 45.2 & 24.9 & \cellcolor[HTML]{FFFFED}38.5 & 38.2 & 28.1 & \cellcolor[HTML]{FFFFED}33.1 \\
PROB~\cite{zohar2023prob} & \cellcolor[HTML]{FFFFED}17.6 & 73.4 & \cellcolor[HTML]{FFFFED}22.3 & 66.3 & 36.0 & \cellcolor[HTML]{FFFFED}50.4 & \cellcolor[HTML]{FFFFED}24.8 & 47.8 & 30.4 & \cellcolor[HTML]{FFFFED}42.0 & 42.6 & 31.7 & \cellcolor[HTML]{FFFFED}39.9 \\
CAT~\cite{ma2023cat} & \cellcolor[HTML]{FFFFED}24.0 & \textbf{74.2} & \cellcolor[HTML]{FFFFED}23.0 & \textbf{67.6} & 35.5 & \cellcolor[HTML]{FFFFED}50.7 & \cellcolor[HTML]{FFFFED}24.6 & 51.2 & 32.6 & \cellcolor[HTML]{FFFFED}45.0 & 45.4 & 35.1 & \cellcolor[HTML]{FFFFED}42.8 \\
OrthorganalDet~\cite{sun2024exploring} & \cellcolor[HTML]{FFFFED}\underline{24.6} & 71.6 & \cellcolor[HTML]{FFFFED}\underline{27.9} & 64.0 & \underline{39.9} & \cellcolor[HTML]{FFFFED}\underline{51.3} & \cellcolor[HTML]{FFFFED}\underline{31.9} & \underline{52.1} & \textbf{42.2} & \cellcolor[HTML]{FFFFED}\underline{48.8} & \textbf{48.7} & \underline{38.8} & \textbf{46.2}\\

\rowcolor{lightyellow}
\textbf{UADet} & \cellcolor[HTML]{FFFFED}\textbf{30.0} & \underline{73.9} & \cellcolor[HTML]{FFFFED}\textbf{36.6} & \underline{66.8} & \textbf{43.0} & \cellcolor[HTML]{FFFFED}\textbf{54.3} & \cellcolor[HTML]{FFFFED}\textbf{38.5} & \textbf{52.8} & \underline{40.6} & \cellcolor[HTML]{FFFFED}\textbf{48.8} & \underline{48.1} & \textbf{39.0} & \cellcolor[HTML]{FFFFED}\underline{46.0} \\
\hline

\end{tabular}%
}
    \label{tab:owod}
\end{table*}

\subsection{Experimental Setup}
\label{subsec:exp_setup}
\noindent {\bf Datasets.}
We evaluate UADet on benchmarks introduced by OpenDet~\cite{han2022expanding}, which is constructed based on PASCAL VOC0712~\cite{everingham2010pascal} and MS COCO~\cite{lin2014microsoft} datasets. The closed-set training is conducted on VOC07 train and VOC12 trainval sets, with closed-set evaluation performed on the VOC07 test split.  For open-set evaluation, 20 VOC classes and 60 non-VOC classes from COCO are utilized, forming two distinct settings: VOC-COCO-\{T$_1$,  T$_2$\}. Furthermore, we extend UADet for OWOD on two benchmarks: the ``superclass-mixed OWOD benchmark'' (M-OWODB) by~\cite{joseph2021towards} and the ``superclass-separated OWOD benchmark'' (S-OWODB) by~\cite{gupta2022ow}. The M-OWODB benchmark includes both COCO and PASCAL VOC, while the S-OWODB benchmark utilizes only COCO, with both grouped into four non-overlapping tasks, each representing an OSOD task. More details on dataset splits are provided in the supplementary materials.

\noindent {\bf Evaluation metrics.} 
We assess closed-set performance using the mean average precision of known classes (mAP). For open-set performance evaluation, we report four metrics: Wilderness Impact (WI), Absolute Open-Set Error (A-OSE), Unknown Average Precision (U-AP), and Unkown Recall (U-Recall) of the unknown classes. Specifically, WI and A-OSE are used to evaluate the model's confusion of unknown objects with known classes. \textit{Note that for OSOD, U-Recall is a more reliable evaluation metric than U-AP.} 
In the OSOD task, objects not labeled by humans can still be detected by an OSOD detector in a fixed-category dataset, causing the standard AP calculation to treat them as false positives instead of true positives. However, U-Recall is unaffected by this issue, making it a reliable metric for the OSOD task. Similar concerns have also been discussed in prior work on open-world detection ~\cite{cao2024hassod, gupta2022ow, zohar2023prob}, segmentation~\cite{wang2022open}, and tracking~\cite{liu2022opening}. Thus, we prioritize the U-Recall metric for the OSOD task. Despite the potential flaw in U-AP, UADet achieves state-of-the-art or comparable performance on both U-AP and U-Recall metrics. We provide a more detailed analysis in the supplementary.

\noindent {\bf Implementation details.} In our framework, we adopt the ImageNet-pretrained ResNet-50~\cite{he2016resnet} with Feature Pyramid Network (FPN)~\cite{lin2017feature} as the backbone for comparison in OSOD and OWOD benchmarks. To verify the generalization ability, we also evaluate our method using a DINO-pretrained ResNet-50 backbone.  We follow the same learning rate schedules as Detectron2 ~\cite{wu2019detectron2}. For optimization, we employ the SGD optimizer with an initial learning rate of 0.01, momentum of 0.9, and weight decay of 0.0001. Training is performed on 4$\times$RTX3090 GPUs with a batch size of 16.


\subsection{Main Results} \label{subsec:}
\noindent {\bf Results on VOC-COCO-\{T$_1$, T$_2$\}.} 
We comprehensively evaluate UADet on VOC-COCO-T$_1$ and VOC-COCO-T$_2$ benchmarks against state-of-the-art OSOD methods, as shown in Tables~\ref{tab:voc_coco_a} and~\ref{tab:voc_coco_b}. With ImageNet pre-trained backbone, UADet significantly outperforms previous methods by achieving an average U-Recall improvement of 20.5\% over OpenDet across six benchmarks while maintaining comparable performance in WI, A-OSE, and mAP. Notably, despite recalling twice as many unknown objects, UADet still achieves 1.1\% higher U-AP than OpenDet averaged across all benchmarks, demonstrating its superior unknown object detection capability. When equipped with DINO self-supervised pre-trained backbone, UADet exhibits even more remarkable performance - it surpasses PROB by a large margin of 25.6\% in U-Recall (62.73\% vs. 37.10\%) on VOC-COCO-20000 while achieving better results in other metrics. The consistent superior performance across different scales of unknown instances (from 2,500 to 20,000) further validates the effectiveness and robustness of our approach.

\noindent\textbf{Results on M-OWODB and S-OWODB.} 
We further evaluate OADet on OWOD benchmarks, with detailed results shown in Table~\ref{tab:owod}. Our method consistently outperforms existing OWOD approaches across different metrics and tasks. On both M-OWODB and S-OWODB, OADet demonstrates substantially higher unknown recall compared to OrthorganalDet, with an average improvement of $13.83\%$ and $6.9\%$ respectively. This significant gain in unknown object detection does not come at the cost of known class performance. Instead, OADet maintains competitive or even better mAP scores for both previously known and currently known categories across all tasks. These results demonstrate that our uncertainty-aware strategy effectively addresses the inherent ambiguity in open-world scenarios, achieving a better balance between unknown object discovery and known object detection.

\subsection{Ablation Study}
\label{subsec:ablation}
To comprehensively investigate our core hypotheses and model designs, we conduct a series of ablative studies on the VOC-COCO-20 benchmark. We adopt ImageNet-pretrained ResNet-50 with FPN as our backbone network and trained with the default settings unless otherwise specified.

\begin{table}[t] 
    \vspace{-2mm}
    \caption{\textbf{Effectiveness of different uncertainty measures on VOC-COCO-20.}}


\centering
\scriptsize
\setlength{\tabcolsep}{1.5mm}
\begin{tabular}{cc|ccccc}
\hline
Geo. & App. & WI ($\downarrow$) & A-OSE ($\downarrow$)  & mAP ($\uparrow$)  & U-AP ($\uparrow$)   & U-Recall ($\uparrow$) \\
\hline
\multicolumn{2}{c|}{baseline} & 19.40 & 17196 & 58.90 & 0 & 0  \\
$\checkmark$ & & 15.31 & 10974 & 59.06 & 0.05 & 16.60 \\
& $\checkmark$ & 13.80 & 11294 & 58.13 & 10.43 & 58.61 \\
$\checkmark$ & $\checkmark$ & \textbf{13.19} & \textbf{10186} & \textbf{59.12} & \textbf{15.09} & \textbf{59.03} \\
\hline
\end{tabular}

    \label{tab:effectiveness_uncertainty_measure}
\end{table}

\begin{table} 
    \caption{\textbf{Exploration of different types of negative proposals on VOC-COCO-20.}}
    \vspace{-3mm}
\begin{center}
\scriptsize
\begin{tabular}{l|ccccc}
\hline
Method            & WI ($\downarrow$) & A-OSE ($\downarrow$)  & mAP ($\uparrow$)  & U-AP ($\uparrow$)   & U-Recall ($\uparrow$) \\
\hline
A & 19.40 & 17196 & 58.90 & 0 & 0  \\
B & {\bf 12.28} & 10847 & 58.90 & 12.85 &  49.51\\
C & 13.54 & 10348 & 59.05 &  14.57 &  54.74\\
D & 13.19 & {\bf 10186} & {\bf 59.12} & {\bf 15.09} & {\bf 59.03}\\
\hline
\end{tabular}%
\end{center}

    \vspace{-6mm}
    \label{tab:explore_neg}
\end{table}

\begin{figure*}[t]
        \centering
        \includegraphics[width=0.8\linewidth]{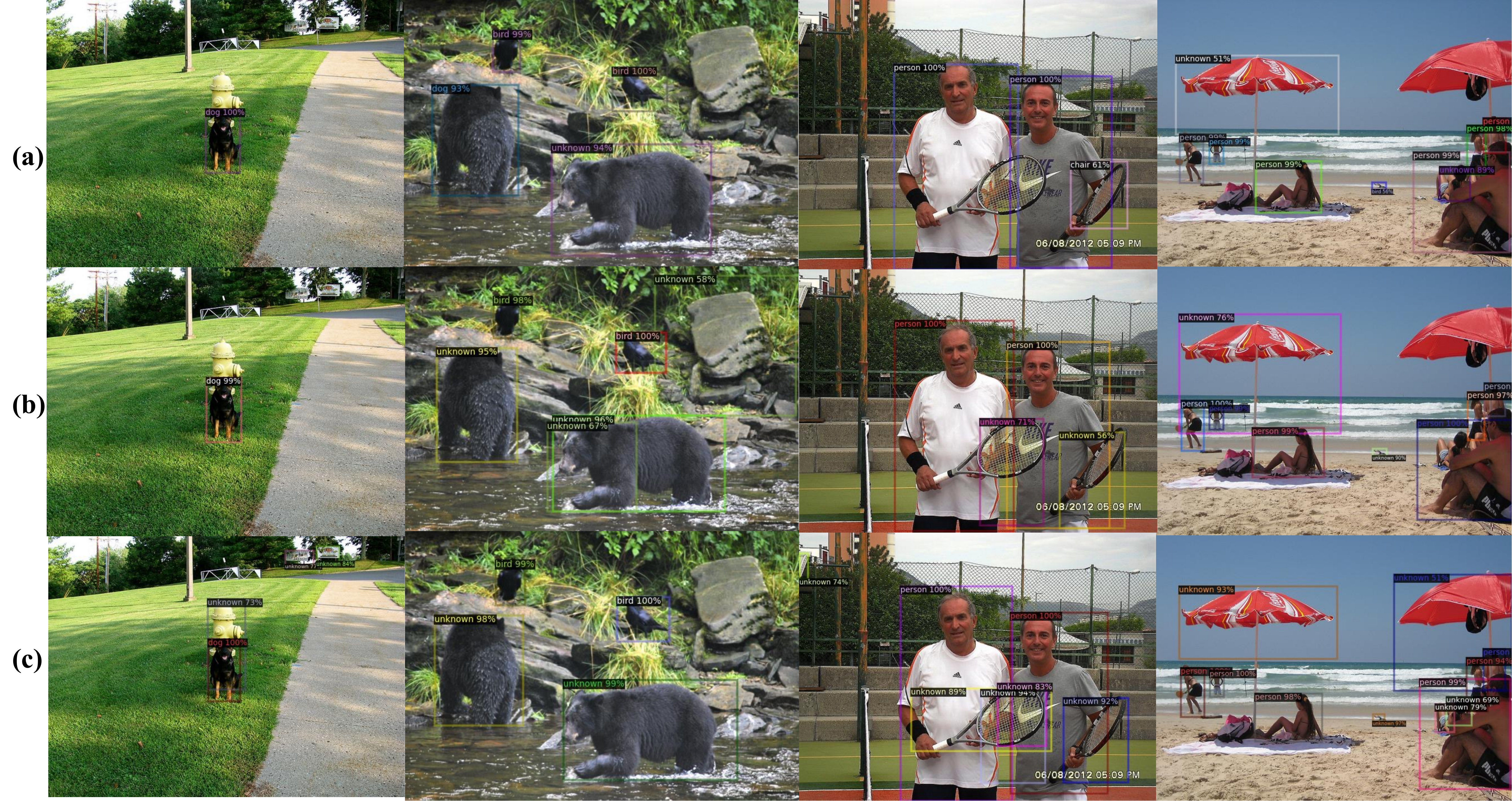}
        \caption{\textbf{Qualitative comparisons between the OpenDet (top), top-\textit{k} hard pseudo-labeling strategy (middle) and ours (bottom).} All models are trained on VOC, and the detection results on COCO are visualized. It is important to note that for improved visualization, we apply NMS between known classes and the unknown class.}
        \label{fig:qualitative_results}
        \vspace{-4mm}
\end{figure*}

\noindent {\bf Effectiveness of different uncertainty measures.} We utilize the standard Faster R-CNN as our baseline model. From the results presented in Table~\ref{tab:effectiveness_uncertainty_measure}, we can observe that relying solely on geometric uncertainty-based pseudo labels leads to numerous low-quality detections of unknown classes, as evidenced by the lower U-AP and U-Recall. This may be due to the model's tendency to mistreat many pure background regions as unknown classes since the assigned pseudo-labels depend solely on the IoU with known objects, as per Eq.~\ref{eq:geo_un}. Conversely, the utilization of appearance uncertainty-based pseudo-labels notably enhances the model's open-set performance, underscoring the reliability of RPN's confidence scores as pseudo-labels for identifying unknown objects. Upon introducing the geometry uncertainty-based pseudo labeling method alongside the appearance uncertainty-based pseudo labeling strategy, we observe an almost 1\% increase in close-set performance, measured by mAP. Additionally, the open-set performance, measured by U-AP and U-Recall, also increases by 4.66\% and 0.42\% respectively. This indicates that the geometry-based score can effectively mitigate the model's confusion between known and unknown classes.

\noindent {\bf Exploration of different types of negative proposals.} 
We evaluate different variants of our approach against the standard Faster R-CNN baseline (model A). Using only geometry uncertainty (model B) demonstrates a clear improvement in open-set detection capability while maintaining the close-set performance, suggesting that geometric relationships between proposals and GT boxes provide valuable signals for unknown object detection. When incorporating appearance uncertainty alone (model C), we observe further improvements in both open-set and close-set metrics, which validates our hypothesis that RPN's objectness scores serve as reliable indicators for unknown object detection. Our full model (D), which combines both uncertainty measures, achieves the best performance across all metrics. Notably, the significant reduction in WI and AOSE scores (compared to baseline) indicates that our dual uncertainty mechanism effectively mitigates the confusion between known and unknown classes. The simultaneous improvements in both mAP and U-AP further demonstrate that our approach successfully maintains known class performance while enhancing unknown object detection, addressing a key challenge in OSOD.

\noindent {\bf Qualitative comparisons.} 
Fig.~\ref{fig:qualitative_results} provides a qualitative comparison between UADet, OpenDet, and the top-\textit{k} hard pseudo-labeling method. Our method demonstrates several key advantages: (1) successful detection of unknown objects that overlap with known objects (e.g., fire hydrants), where other methods fail to detect; (2) more accurate localization of unknown objects (e.g., complete tennis rackets rather than partial detections), while OpenDet misses these objects entirely and the top-\textit{k} method only captures partial regions; and (3) better robustness to background noise, effectively avoiding false detections in complex backgrounds that often mislead the top-\textit{k} method. These qualitative results align with our quantitative findings and further validate the effectiveness of our uncertainty-aware strategy.


\section{Conclusion}
This paper proposes a simple yet effective uncertainty-aware framework UADet, which introduces uncertainty awareness by considering both appearance uncertainty and geometry uncertainty, enabling the framework to properly make use of unannotated instances to facilitate unknown object detection. Despite its simplicity, UADet demonstrates superior performance through extensive experiments. On OSOD benchmarks, it achieves a 1.8× increase in unknown recall while maintaining high performance in known object detection. When extended to the more challenging OWOD scenario, UADet significantly outperforms the previous SOTA, achieving average improvements of 13.8\% and 6.9\% in unknown recall on M-OWODB and S-OWODB benchmarks respectively, resulting in an overall 10.4\% gain. These results comprehensively validate the effectiveness of our uncertainty-aware strategy across different open-set scenarios.

\section*{Acknowledgement}
This work is supported by Hong Kong Research Grant Council - Early Career Scheme (Grant No. 27208022), National Natural Science Foundation of China (Grant No. 62306251),
and HKU Seed Fund for Basic Research.
{
    \small
    \bibliographystyle{ieeenat_fullname}
    \bibliography{main}

\begin{thebibliography}{57}
\providecommand{\natexlab}[1]{#1}
\providecommand{\url}[1]{\texttt{#1}}
\expandafter\ifx\csname urlstyle\endcsname\relax
  \providecommand{\doi}[1]{doi: #1}\else
  \providecommand{\doi}{doi: \begingroup \urlstyle{rm}\Url}\fi

\bibitem[Bendale and Boult(2016)]{bendale2016towards}
Abhijit Bendale and Terrance~E Boult.
\newblock Towards open set deep networks.
\newblock In \emph{CVPR}, 2016.

\bibitem[Cai and Vasconcelos(2018)]{cai2018cascade}
Zhaowei Cai and Nuno Vasconcelos.
\newblock Cascade r-cnn: Delving into high quality object detection.
\newblock In \emph{CVPR}, 2018.

\bibitem[Cao et~al.(2024)Cao, Joshi, Gui, and Wang]{cao2024hassod}
Shengcao Cao, Dhiraj Joshi, Liangyan Gui, and Yu-Xiong Wang.
\newblock Hassod: Hierarchical adaptive self-supervised object detection.
\newblock In \emph{NeurIPS}, 2024.

\bibitem[Carion et~al.(2020)Carion, Massa, Synnaeve, Usunier, Kirillov, and Zagoruyko]{carion2020end}
Nicolas Carion, Francisco Massa, Gabriel Synnaeve, Nicolas Usunier, Alexander Kirillov, and Sergey Zagoruyko.
\newblock End-to-end object detection with transformers.
\newblock In \emph{ECCV}, 2020.

\bibitem[Chen et~al.(2020)Chen, Qiao, Shi, Peng, Li, Huang, Pu, and Tian]{chen2020learning}
Guangyao Chen, Limeng Qiao, Yemin Shi, Peixi Peng, Jia Li, Tiejun Huang, Shiliang Pu, and Yonghong Tian.
\newblock Learning open set network with discriminative reciprocal points.
\newblock In \emph{ECCV}. Springer, 2020.

\bibitem[Chen et~al.(2021)Chen, Peng, Wang, and Tian]{chen2021adversarial}
Guangyao Chen, Peixi Peng, Xiangqian Wang, and Yonghong Tian.
\newblock Adversarial reciprocal points learning for open set recognition.
\newblock \emph{arXiv preprint arXiv:2103.00953}, 2021.

\bibitem[Dhamija et~al.(2020)Dhamija, Gunther, Ventura, and Boult]{dhamija2020overlooked}
Akshay Dhamija, Manuel Gunther, Jonathan Ventura, and Terrance Boult.
\newblock The overlooked elephant of object detection: Open set.
\newblock In \emph{WACV}, 2020.

\bibitem[Everingham et~al.(2010)Everingham, Van~Gool, Williams, Winn, and Zisserman]{everingham2010pascal}
Mark Everingham, Luc Van~Gool, Christopher~KI Williams, John Winn, and Andrew Zisserman.
\newblock The pascal visual object classes (voc) challenge.
\newblock \emph{IJCV}, 2010.

\bibitem[Gal and Ghahramani(2016)]{gal2016dropout}
Yarin Gal and Zoubin Ghahramani.
\newblock Dropout as a bayesian approximation: Representing model uncertainty in deep learning.
\newblock In \emph{ICML}, 2016.

\bibitem[Ge et~al.(2017)Ge, Demyanov, Chen, and Garnavi]{ge2017generative}
ZongYuan Ge, Sergey Demyanov, Zetao Chen, and Rahil Garnavi.
\newblock Generative openmax for multi-class open set classification.
\newblock In \emph{BMVC}, 2017.

\bibitem[Girshick(2015)]{girshick2015fast}
Ross Girshick.
\newblock {Fast R-CNN}.
\newblock In \emph{ICCV}, 2015.

\bibitem[Gupta et~al.(2022)Gupta, Narayan, Joseph, Khan, Khan, and Shah]{gupta2022ow}
Akshita Gupta, Sanath Narayan, KJ Joseph, Salman Khan, Fahad~Shahbaz Khan, and Mubarak Shah.
\newblock Ow-detr: Open-world detection transformer.
\newblock In \emph{CVPR}, 2022.

\bibitem[Han et~al.(2022)Han, Ren, Ding, Pan, Yan, and Xia]{han2022expanding}
Jiaming Han, Yuqiang Ren, Jian Ding, Xingjia Pan, Ke Yan, and Gui-Song Xia.
\newblock Expanding low-density latent regions for open-set object detection.
\newblock In \emph{CVPR}, 2022.

\bibitem[He et~al.(2016)He, Zhang, Ren, and Sun]{he2016resnet}
Kaiming He, Xiangyu Zhang, Shaoqing Ren, and Jian Sun.
\newblock Deep residual learning for image recognition.
\newblock In \emph{CVPR}, 2016.

\bibitem[He et~al.(2017)He, Gkioxari, Doll{\'a}r, and Girshick]{he2017mask}
Kaiming He, Georgia Gkioxari, Piotr Doll{\'a}r, and Ross Girshick.
\newblock Mask r-cnn.
\newblock In \emph{ICCV}, 2017.

\bibitem[Jain et~al.(2014)Jain, Scheirer, and Boult]{jain2014multi}
Lalit~P Jain, Walter~J Scheirer, and Terrance~E Boult.
\newblock Multi-class open set recognition using probability of inclusion.
\newblock In \emph{ECCV}, 2014.

\bibitem[Joseph et~al.(2021)Joseph, Khan, Khan, and Balasubramanian]{joseph2021towards}
KJ Joseph, Salman Khan, Fahad~Shahbaz Khan, and Vineeth~N Balasubramanian.
\newblock Towards open world object detection.
\newblock In \emph{CVPR}, 2021.

\bibitem[J{\'u}nior et~al.(2017)J{\'u}nior, De~Souza, Werneck, Stein, Pazinato, de~Almeida, Penatti, Torres, and Rocha]{junior2017nearest}
Pedro R~Mendes J{\'u}nior, Roberto~M De~Souza, Rafael de~O Werneck, Bernardo~V Stein, Daniel~V Pazinato, Waldir~R de Almeida, Ot{\'a}vio~AB Penatti, Ricardo da~S Torres, and Anderson Rocha.
\newblock Nearest neighbors distance ratio open-set classifier.
\newblock \emph{Machine Learning}, 2017.

\bibitem[Kuznetsova et~al.(2020)Kuznetsova, Rom, Alldrin, Uijlings, Krasin, Pont-Tuset, Kamali, Popov, Malloci, Kolesnikov, et~al.]{kuznetsova2020open}
Alina Kuznetsova, Hassan Rom, Neil Alldrin, Jasper Uijlings, Ivan Krasin, Jordi Pont-Tuset, Shahab Kamali, Stefan Popov, Matteo Malloci, Alexander Kolesnikov, et~al.
\newblock The open images dataset v4: Unified image classification, object detection, and visual relationship detection at scale.
\newblock \emph{IJCV}, 2020.

\bibitem[Li et~al.(2022)Li, Dai, Ma, Liu, Chen, Wu, He, Kitani, and Vajda]{li2022cross}
Yu-Jhe Li, Xiaoliang Dai, Chih-Yao Ma, Yen-Cheng Liu, Kan Chen, Bichen Wu, Zijian He, Kris Kitani, and Peter Vajda.
\newblock Cross-domain adaptive teacher for object detection.
\newblock In \emph{CVPR}, 2022.

\bibitem[Lin et~al.(2014)Lin, Maire, Belongie, Hays, Perona, Ramanan, Doll{\'a}r, and Zitnick]{lin2014microsoft}
Tsung-Yi Lin, Michael Maire, Serge Belongie, James Hays, Pietro Perona, Deva Ramanan, Piotr Doll{\'a}r, and C~Lawrence Zitnick.
\newblock Microsoft coco: Common objects in context.
\newblock In \emph{ECCV}, 2014.

\bibitem[Lin et~al.(2017{\natexlab{a}})Lin, Doll{\'a}r, Girshick, He, Hariharan, and Belongie]{lin2017feature}
Tsung-Yi Lin, Piotr Doll{\'a}r, Ross Girshick, Kaiming He, Bharath Hariharan, and Serge Belongie.
\newblock Feature pyramid networks for object detection.
\newblock In \emph{CVPR}, 2017{\natexlab{a}}.

\bibitem[Lin et~al.(2017{\natexlab{b}})Lin, Goyal, Girshick, He, and Doll{\'a}r]{lin2017focal}
Tsung-Yi Lin, Priya Goyal, Ross Girshick, Kaiming He, and Piotr Doll{\'a}r.
\newblock Focal loss for dense object detection.
\newblock In \emph{ICCV}, 2017{\natexlab{b}}.

\bibitem[Liu et~al.(2016)Liu, Anguelov, Erhan, Szegedy, Reed, Fu, and Berg]{liu2016ssd}
Wei Liu, Dragomir Anguelov, Dumitru Erhan, Christian Szegedy, Scott Reed, Cheng-Yang Fu, and Alexander~C Berg.
\newblock {SSD}: Single shot multibox detector.
\newblock In \emph{ECCV}, 2016.

\bibitem[Liu et~al.(2022)Liu, Zulfikar, Luiten, Dave, Ramanan, Leibe, O{\v{s}}ep, and Leal-Taix{\'e}]{liu2022opening}
Yang Liu, Idil~Esen Zulfikar, Jonathon Luiten, Achal Dave, Deva Ramanan, Bastian Leibe, Aljo{\v{s}}a O{\v{s}}ep, and Laura Leal-Taix{\'e}.
\newblock Opening up open world tracking.
\newblock In \emph{CVPR}, pages 19045--19055, 2022.

\bibitem[Liu et~al.(2021)Liu, Lin, Cao, Hu, Wei, Zhang, Lin, and Guo]{liu2021swin}
Ze Liu, Yutong Lin, Yue Cao, Han Hu, Yixuan Wei, Zheng Zhang, Stephen Lin, and Baining Guo.
\newblock Swin transformer: Hierarchical vision transformer using shifted windows.
\newblock In \emph{CVPR}, 2021.

\bibitem[Ma et~al.(2023{\natexlab{a}})Ma, Wang, Wei, Fan, Li, Liu, and Lv]{ma2023cat}
Shuailei Ma, Yuefeng Wang, Ying Wei, Jiaqi Fan, Thomas~H Li, Hongli Liu, and Fanbing Lv.
\newblock Cat: Localization and identification cascade detection transformer for open-world object detection.
\newblock In \emph{CVPR}, 2023{\natexlab{a}}.

\bibitem[Ma et~al.(2023{\natexlab{b}})Ma, Li, Zhang, Guo, Zhang, Gong, and Liu]{ma2023annealing}
Yuqing Ma, Hainan Li, Zhange Zhang, Jinyang Guo, Shanghang Zhang, Ruihao Gong, and Xianglong Liu.
\newblock Annealing-based label-transfer learning for open world object detection.
\newblock In \emph{CVPR}, 2023{\natexlab{b}}.

\bibitem[Miller et~al.(2018)Miller, Nicholson, Dayoub, and S{\"u}nderhauf]{miller2018dropout}
Dimity Miller, Lachlan Nicholson, Feras Dayoub, and Niko S{\"u}nderhauf.
\newblock Dropout sampling for robust object detection in open-set conditions.
\newblock In \emph{ICRA}, 2018.

\bibitem[Miller et~al.(2019)Miller, Dayoub, Milford, and S{\"u}nderhauf]{miller2019evaluating}
Dimity Miller, Feras Dayoub, Michael Milford, and Niko S{\"u}nderhauf.
\newblock Evaluating merging strategies for sampling-based uncertainty techniques in object detection.
\newblock In \emph{ICRA}, 2019.

\bibitem[Miller et~al.(2021)Miller, S{\"u}nderhauf, Milford, and Dayoub]{miller2021uncertainty}
Dimity Miller, Niko S{\"u}nderhauf, Michael Milford, and Feras Dayoub.
\newblock Uncertainty for identifying open-set errors in visual object detection.
\newblock \emph{ICRA}, 2021.

\bibitem[Neal et~al.(2018)Neal, Olson, Fern, Wong, and Li]{neal2018open}
Lawrence Neal, Matthew Olson, Xiaoli Fern, Weng-Keen Wong, and Fuxin Li.
\newblock Open set learning with counterfactual images.
\newblock In \emph{ECCV}, 2018.

\bibitem[Oza and Patel(2019)]{oza2019c2ae}
Poojan Oza and Vishal~M Patel.
\newblock C2ae: Class conditioned auto-encoder for open-set recognition.
\newblock In \emph{CVPR}, 2019.

\bibitem[Redmon et~al.(2016)Redmon, Divvala, Girshick, and Farhadi]{redmon2016you}
Joseph Redmon, Santosh Divvala, Ross Girshick, and Ali Farhadi.
\newblock You only look once: Unified, real-time object detection.
\newblock In \emph{CVPR}, 2016.

\bibitem[Ren et~al.(2017)Ren, He, Girshick, and Sun]{ren2017faster}
Shaoqing Ren, Kaiming He, Ross Girshick, and Jian Sun.
\newblock {Faster R-CNN}: Towards real-time object detection with region proposal networks.
\newblock \emph{IEEE TPAMI}, 2017.

\bibitem[Scheirer et~al.(2012)Scheirer, de~Rezende~Rocha, Sapkota, and Boult]{scheirer2012toward}
Walter~J Scheirer, Anderson de Rezende~Rocha, Archana Sapkota, and Terrance~E Boult.
\newblock Toward open set recognition.
\newblock \emph{IEEE TPAMI}, 2012.

\bibitem[Scheirer et~al.(2014)Scheirer, Jain, and Boult]{scheirer2014probability}
Walter~J Scheirer, Lalit~P Jain, and Terrance~E Boult.
\newblock Probability models for open set recognition.
\newblock \emph{IEEE TPAMI}, 2014.

\bibitem[Shao et~al.(2019)Shao, Li, Zhang, Peng, Yu, Zhang, Li, and Sun]{shao2019objects365}
Shuai Shao, Zeming Li, Tianyuan Zhang, Chao Peng, Gang Yu, Xiangyu Zhang, Jing Li, and Jian Sun.
\newblock Objects365: A large-scale, high-quality dataset for object detection.
\newblock In \emph{CVPR}, 2019.

\bibitem[Sun et~al.(2020)Sun, Yang, Zhang, Ling, and Peng]{sun2020conditional}
Xin Sun, Zhenning Yang, Chi Zhang, Keck-Voon Ling, and Guohao Peng.
\newblock Conditional gaussian distribution learning for open set recognition.
\newblock In \emph{CVPR}, 2020.

\bibitem[Sun et~al.(2024)Sun, Li, and Mu]{sun2024exploring}
Zhicheng Sun, Jinghan Li, and Yadong Mu.
\newblock Exploring orthogonality in open world object detection.
\newblock In \emph{CVPR}, pages 17302--17312, 2024.

\bibitem[Tan et~al.(2020)Tan, Wang, Li, Li, Ouyang, Yin, and Yan]{tan2020equalization}
Jingru Tan, Changbao Wang, Buyu Li, Quanquan Li, Wanli Ouyang, Changqing Yin, and Junjie Yan.
\newblock Equalization loss for long-tailed object recognition.
\newblock In \emph{CVPR}, 2020.

\bibitem[Tian et~al.(2019)Tian, Shen, Chen, and He]{tian2019fcos}
Zhi Tian, Chunhua Shen, Hao Chen, and Tong He.
\newblock Fcos: Fully convolutional one-stage object detection.
\newblock In \emph{ICCV}, 2019.

\bibitem[Wang et~al.(2022)Wang, Feiszli, Wang, Malik, and Tran]{wang2022open}
Weiyao Wang, Matt Feiszli, Heng Wang, Jitendra Malik, and Du Tran.
\newblock Open-world instance segmentation: Exploiting pseudo ground truth from learned pairwise affinity.
\newblock In \emph{CVPR}, 2022.

\bibitem[Wang et~al.(2020)Wang, Huang, Darrell, Gonzalez, and Yu]{wang2020frustratingly}
Xin Wang, Thomas~E Huang, Trevor Darrell, Joseph~E Gonzalez, and Fisher Yu.
\newblock Frustratingly simple few-shot object detection.
\newblock In \emph{ICML}, 2020.

\bibitem[Wang et~al.(2023)Wang, Yue, Hua, and Zhang]{wang2023random}
Yanghao Wang, Zhongqi Yue, Xian-Sheng Hua, and Hanwang Zhang.
\newblock Random boxes are open-world object detectors.
\newblock In \emph{ICCV}, 2023.

\bibitem[Wu et~al.(2019)Wu, Kirillov, Massa, Lo, and Girshick]{wu2019detectron2}
Yuxin Wu, Alexander Kirillov, Francisco Massa, Wan-Yen Lo, and Ross Girshick.
\newblock Detectron2.
\newblock \url{https://github.com/facebookresearch/detectron2}, 2019.

\bibitem[Wu et~al.(2022{\natexlab{a}})Wu, Zhao, Ma, Wang, and Liu]{wu2022two}
Yan Wu, Xiaowei Zhao, Yuqing Ma, Duorui Wang, and Xianglong Liu.
\newblock Two-branch objectness-centric open world detection.
\newblock In \emph{HCMA}, 2022{\natexlab{a}}.

\bibitem[Wu et~al.(2022{\natexlab{b}})Wu, Lu, Chen, Wu, Kang, and Yu]{wu2022uc}
Zhiheng Wu, Yue Lu, Xingyu Chen, Zhengxing Wu, Liwen Kang, and Junzhi Yu.
\newblock Uc-owod: Unknown-classified open world object detection.
\newblock In \emph{ECCV}, 2022{\natexlab{b}}.

\bibitem[Xu et~al.(2021)Xu, Zhang, Hu, Wang, Wang, Wei, Bai, and Liu]{xu2021end}
Mengde Xu, Zheng Zhang, Han Hu, Jianfeng Wang, Lijuan Wang, Fangyun Wei, Xiang Bai, and Zicheng Liu.
\newblock End-to-end semi-supervised object detection with soft teacher.
\newblock In \emph{ICCV}, 2021.

\bibitem[Yang et~al.(2021)Yang, Sun, Jiang, Xia, Zhang, Yuan, Wang, Luo, and Xu]{yang2021objects}
Shuo Yang, Peize Sun, Yi Jiang, Xiaobo Xia, Ruiheng Zhang, Zehuan Yuan, Changhu Wang, Ping Luo, and Min Xu.
\newblock Objects in semantic topology.
\newblock In \emph{ICLR}, 2021.

\bibitem[Yoshihashi et~al.(2019)Yoshihashi, Shao, Kawakami, You, Iida, and Naemura]{yoshihashi2019classification}
Ryota Yoshihashi, Wen Shao, Rei Kawakami, Shaodi You, Makoto Iida, and Takeshi Naemura.
\newblock Classification-reconstruction learning for open-set recognition.
\newblock In \emph{CVPR}, 2019.

\bibitem[Yu et~al.(2023)Yu, Ma, Li, Peng, and Xie]{yu2023open}
Jinan Yu, Liyan Ma, Zhenglin Li, Yan Peng, and Shaorong Xie.
\newblock Open-world object detection via discriminative class prototype learning.
\newblock \emph{arXiv preprint arXiv:2302.11757}, 2023.

\bibitem[Zhang and Patel(2016)]{zhang2016sparse}
He Zhang and Vishal~M Patel.
\newblock Sparse representation-based open set recognition.
\newblock \emph{IEEE TPAMI}, 2016.

\bibitem[Zheng et~al.(2022)Zheng, Li, Hong, Petersson, and Barnes]{zheng2022towards}
Jiyang Zheng, Weihao Li, Jie Hong, Lars Petersson, and Nick Barnes.
\newblock Towards open-set object detection and discovery.
\newblock In \emph{CVPR}, 2022.

\bibitem[Zhou et~al.(2021)Zhou, Ye, and Zhan]{zhou2021learning}
Da-Wei Zhou, Han-Jia Ye, and De-Chuan Zhan.
\newblock Learning placeholders for open-set recognition.
\newblock In \emph{CVPR}, 2021.

\bibitem[Zhu et~al.(2020)Zhu, Su, Lu, Li, Wang, and Dai]{zhu2020deformable}
Xizhou Zhu, Weijie Su, Lewei Lu, Bin Li, Xiaogang Wang, and Jifeng Dai.
\newblock Deformable detr: Deformable transformers for end-to-end object detection.
\newblock In \emph{ICLR}, 2020.

\bibitem[Zohar et~al.(2023)Zohar, Wang, and Yeung]{zohar2023prob}
Orr Zohar, Kuan-Chieh Wang, and Serena Yeung.
\newblock Prob: Probabilistic objectness for open world object detection.
\newblock In \emph{CVPR}, 2023.

\end{thebibliography}
}

\clearpage
\setcounter{page}{1}
\maketitlesupplementary
\setcounter{section}{0}  
\setcounter{table}{0}    
\setcounter{figure}{0}   

\renewcommand{\thesection}{S\arabic{section}}
\renewcommand{\thetable}{S\arabic{table}}
\renewcommand{\thefigure}{S\arabic{figure}} 

In this supplementary material, we present additional quantitative and qualitative results. Section ~\ref{sec:disscusion_unk_ap}, we discuss the limitation of U-AP compared to U-R. Section~\ref{sec:more_exp_details} provides detailed information about the OSOD and OWOD benchmark, evaluation metrics, and implementation details. In Section~\ref{sec:main_results_add}, we provide UADet's performance when leveraging Swin Transformer as the backbone network.
In Section~\ref{sec:ablation_add}, we conduct experiments to analyze the impact of different combinations of appearance uncertainty and geometry uncertainty, as well as the influence of different backbones. 


\section{Unknown Recall \vs Unknown AP}
\label{sec:disscusion_unk_ap}

\begin{figure}[h]
        \centering
        \includegraphics[width=1.\linewidth]{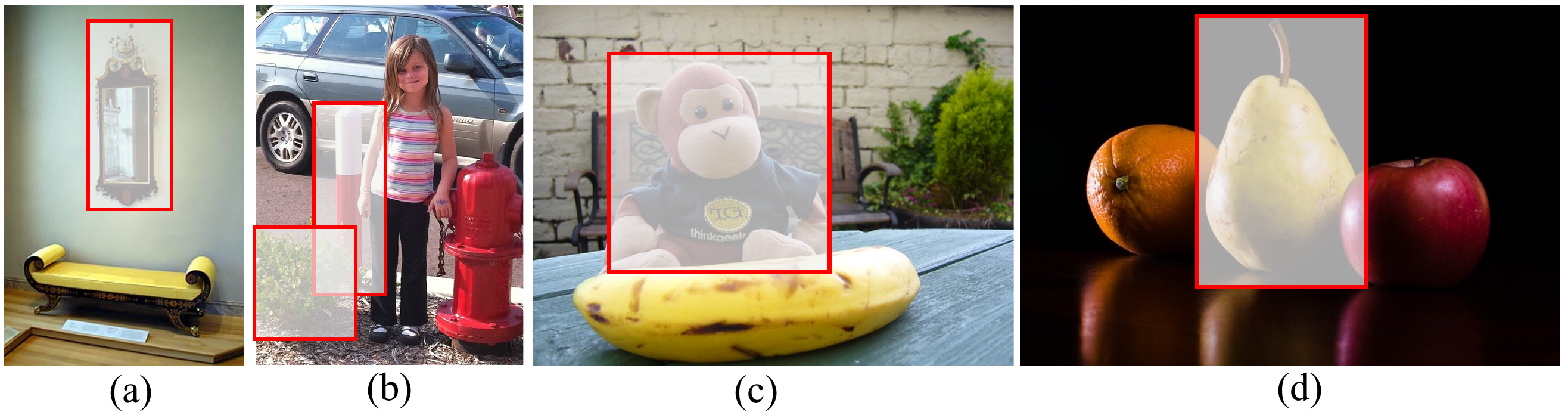}
        \caption{\textbf{Missing annotations in the COCO dataset.} Red boxes denote unlabeled foreground objects in the COCO dataset.}
        \label{fig:Discussion_AP}
\end{figure}


Here, we discuss the limitation of Unknown AP~(U-AP) in measuring the OSOD performance and highlight the importance of Unknown Recal~(U-R).
Particularly, for U-AP, we measure the Precision by $P_u = \frac{TP_u}{TP_u + FP_u}$. The false positive rate $FP_u$ is of concern. Objects from unknown classes may go unannotated in the testing data, as illustrated in Fig.~\ref{fig:Discussion_AP}. Consequently, these objects are inaccurately regarded as false positives in the U-AP calculation, although they should be regarded as true positives (because they are in fact foreground objects). This will lead to a flawed measurement by U-AP. Similar concerns have also been discussed in prior work on open-world detection ~\cite{cao2024hassod, gupta2022ow, zohar2023prob},
segmentation~\cite{wang2022open}, and tracking~\cite{liu2022opening}. In contrast, the U-R will properly consider all the labeled unknown classes without mismeasuring the unlabeled unknown objects. Hence, we regard it as a more convincing metric, despite that our method can achieve superior performance on both metrics.

\section{Experimental Details}
\label{sec:more_exp_details}
\subsection{Datasets}
\noindent {\bf Details about VOC-COCO-\{{T$_{1}$}, {T$_{2}$}\}.} The proposed method is evaluated on benchmarks introduced by OpenDet~\cite{han2022expanding}. OpenDet combines the PASCAL VOC0712 dataset~\cite{everingham2010pascal} with the MS COCO dataset~\cite{lin2014microsoft} to create the OSOD benchmark. Close-set training is performed on the VOC07 \texttt{train} and VOC12 \texttt{trainval} splits. The evaluation is performed under different open-set conditions, with 20 VOC classes and 60 non-VOC classes selected from the COCO dataset. Two settings, VOC-COCO-\{T$_{1}$, T$_{2}$\}, are defined.

In T$_{1}$, three dataset splits (VOC-COCO-\{20, 40, 60\}) are constructed using $n$=5000 VOC testing images, where the number of COCO images increases gradually ($n$, $2n$, $3n$) and contains 20, 40, and 60 non-VOC classes, respectively. The 80 COCO classes are divided into four semantic groups: \textbf{(1)} VOC classes; \textbf{(2)} Outdoor, Accessories, Appliance, Truck; \textbf{(3)} Sports, Food; and \textbf{(4)} Electronic, Indoor, Kitchen, Furniture. Each COCO image contains objects from the corresponding open-set classes but may also include objects from VOC classes, introducing semantic shifts across the datasets.

In T$_{2}$, four dataset splits are created by gradually increasing the Wilderness Ratio (WR), a metric that quantifies the proportion of images containing unknown objects relative to those with known objects. These datasets consist of 5000 VOC testing images and varying numbers of COCO images (2500, 5000, 10000, 20000), which are disjoint from VOC classes. Different form T$_{1}$, T$_{2}$ is designed to evaluate the open-set performance of the model under a higher wilderness condition, where a large number of testing instances are not encountered during training.

\begin{table}[t]
    \caption{{\bf Task composition and statistics in M-OWODB (top) and S-OWODB (bottom) benchmarks.} For each task, we show its semantic categories and the corresponding numbers of training/test images and object instances.}

\centering
\footnotesize
\setlength{\tabcolsep}{4.5pt}
\begin{tabular}{lcccc}
\hline
\textbf{Task IDs ($\rightarrow$)} & \textbf{Task 1} & \textbf{Task 2} & \textbf{Task 3} & \textbf{Task 4} \\
\hline
M-OWODB & \makecell{VOC\\Classes} & \makecell{Outdoor,\\Accessories,\\Appliances,\\Truck} & \makecell{Sports,\\Food} & \makecell{Electronic,\\Indoor,\\Kitchen,\\Furniture} \\
\hline
\# classes & 20 & 20 & 20 & 20 \\
\# training images & 16551 & 45520 & 39402 & 40260 \\
\# test images & 4952 & 1914 & 1642 & 1738 \\
\# training instances & 47223 & 113741 & 114452 & 138996 \\
\# test instances & 14976 & 4966 & 4826 & 6039 \\
\hline
\hline
S-OWODB  & \makecell{Animals,\\Person,\\Vehicles} & \makecell{Outdoor,\\Accessories,\\Appliances,\\Furniture} & \makecell{Sports,\\Food} & \makecell{Electronic,\\Indoor,\\Kitchen} \\
\hline
\# classes & 19 & 21 & 20 & 20 \\
\# training images & 89490 & 55870 & 39402 & 38903 \\
\# test images & 3793 & 2351 & 1642 & 1691 \\
\# training instances & 421243 & 163512 & 114452 & 160794 \\
\# test instances & 17786 & 7159 & 4826 & 7010 \\
\hline
\end{tabular}

    \vspace{-4mm}
\label{tab:benchmark_summary}
\end{table}

\noindent {\bf Details about M-OWODB and S-OWODB.} For open-world object detection, we adopt the superclass-mixed benchmark (M-OWODB) \cite{joseph2021towards} and the superclass-separated benchmark (S-OWODB) \cite{gupta2022ow}. Both benchmarks encompass 80 classes, organized into four sequential tasks, as summarized in Table~\ref{tab:benchmark_summary}.  Specifically, M-OWODB is constructed using the COCO \cite{lin2014microsoft} and PASCAL VOC \cite{everingham2010pascal} datasets. In this benchmark, all classes and data from PASCAL VOC are assigned to the first task, while the remaining COCO classes are distributed across the subsequent three tasks. However, this configuration may result in data leakage across super-classes. For example, although most vehicle-related classes are included in the first task, the truck class is introduced in the second task. To address this issue, S-OWODB employs a more rigorous partitioning of the COCO dataset, ensuring that super-classes are mutually exclusive across tasks. This stricter separation prevents overlap between tasks, thereby facilitating a fairer and more reliable evaluation in the open-world setting.

\subsection{Metrics}
The evaluation of our approach aligns with previous works~\cite{han2022expanding, joseph2021towards, gupta2022ow}, which employ common evaluation metrics. The main metrics utilized are mAP and U-R at an IoU threshold of 0.5. Additionally, we adopt WI at an IoU threshold of 0.8 and A-OSE at an IoU threshold of 0.5 to assess unknown class confusion. Specifically, WI quantifies the degree of misclassification of unknown objects as known classes and is calculated using the formula:
\begin{equation}\label{eq:wi}
    WI = \left(\frac{P_\mathcal{K}}{P_{\mathcal{K} \cup \mathcal{U}}} - 1\right) \times 100,
\end{equation}
where $P_\mathcal{K}$ and $P_{\mathcal{K} \cup \mathcal{U}}$ denote the precision of closed-set classes and open-set classes, respectively. Following~\cite{han2022expanding}, we scale WI by 100 and report it at a recall level of 0.8. As for A-OSE, it indicates the absolute number of unknown objects misclassified as known objects, offering insights into analyzing the confusion of unknown objects.

\subsection{Implementation Details}

\noindent {\bf Detector architecture.}
Following OpenDet~\cite{han2022expanding}, we make three key modifications to the standard Faster R-CNN~\cite{ren2017faster} in its second R-CNN stage.
{\bf (1)} Firstly, we replace the shared fully connected (FC) layer with two parallel FC layers, ensuring that the detection branch and classification branch do not affect each other. {\bf (2)} We adopt a cosine similarity-based classifier rather than a dot-product classifier to reduce intra-class variance~\cite{chen2020learning,wang2020frustratingly}.  {\bf (3)} We make the box regressor class-agnostic, resulting in a length-4 regression output instead of $4(K+2)$, where $K$ represents the number of classes. 

\noindent {\bf Training and inference details.}
Our models are trained using the 3x schedule (\emph{e.g.}, 36 epochs) in line with OpenDet. Additionally, we incorporate the proposed uncertainty-aware pseudo labeling strategy into the training process, starting after a few warmup iterations (\emph{e.g.}, 1000 iterations) to ensure the model generates reliable pseudo labels. During inference, any proposals with objectness scores less than 0.05 are filtered out. Next, the filtered bounding boxes of known and unknown classes are then separately filtered through NMS and the top 100 bounding boxes with the highest scores will be selected as the final prediction results. Within these predictions, we will first keep the top-scoring prediction boxes of known classes, and then select the top-scoring unknown predictions so that the total number of prediction boxes is 100.

\section{Results on Transformer-based Architecture and on VOC-COCO-10000 Dataset}
\label{sec:main_results_add}

To evaluate the generalization capabilities of our proposed UADet, we conduct comprehensive experiments with more advanced backbone architectures. Specifically, we extend our study to transformer-based models, with particular emphasis on the Swin Transformer architecture~\cite{liu2021swin}. The Swin Transformer, known for its hierarchical feature representation and shifted window attention mechanism, serves as a stronger backbone compared to traditional convolutional neural networks. The detailed experimental results and in-depth analysis of these transformer-based implementations are presented in Section~\ref{sec:comparison_swin}. Furthermore, we present additional experimental results on VOC-COCO-10000, with detailed analysis provided in Section~\ref{sec:voc_coco_10000}.

\subsection{Performance analysis with Transformer-based backbones}
\label{sec:comparison_swin}

In Table~\ref{tab:Comparisons_under_transformer_based_backbone}, we compare UADet and OpenDet using ImageNet-pretrained ResNet50 and Swin Transformer backbones. The results reveal two key insights: (1) With the same backbone, UADet outperforms OpenDet on both known and unknown classes. (2) Using a stronger backbone enhances UADet's performance on both known and unknown classes, highlighting its generalization capability.

\begin{table}[t]
    \caption{{\bf Comparisons with other methods on VOC-COCO-20 under transformer-based backbone, \emph{i.e.}, Swin-T~\cite{liu2021swin}}.}

\centering
\small
\setlength{\tabcolsep}{1.1mm}{
\begin{tabular}{c|c|ccc}
\midrule
Backbone & Method & U-Recall ($\uparrow$) & mAP ($\uparrow$) & U-AP ($\uparrow$) \\ \midrule
\multirow{2}{*}{ResNet50~\cite{he2016resnet}} 
& OpenDet~\cite{han2022expanding} & 37.65 & 58.64 & 14.38 \\
& \bf{UADet} & \bf{59.03} & \bf{59.12} & \bf{15.09} \\
\hline
\multirow{2}{*}{Swin-T~\cite{liu2021swin}} 
& OpenDet~\cite{han2022expanding} & 38.90 & 63.42 & 16.35 \\
& \bf{UADet} & \bf{61.74} & \bf{63.95} & \bf{17.53} \\
\midrule
\end{tabular}%
}
    \vspace{-4mm}
\label{tab:Comparisons_under_transformer_based_backbone}
\end{table}

\subsection{Result comparison on VOC-COCO-10000}
\label{sec:voc_coco_10000}
We present additional results on  VOC-COCO-10000 dataset in Table~\ref{tab:Comparisons_with_other methods_on_VOC-COCO-10000}, which highlights the promising open-set performance of our method compared to other methods.

\begin{table}[t]
    \caption{{\bf Comparisons with other methods on VOC-COCO-10000.} The underline denotes the second-best method.}

\centering
\small
\setlength{\tabcolsep}{0.6mm}{
\begin{tabular}{c|ccccc}
\midrule
Method & U-Recall ($\uparrow$) & mAP ($\uparrow$) & U-AP ($\uparrow$) & WI ($\downarrow$) & AOSE ($\downarrow$) \\ 
\midrule
FR-CNN~\cite{ren2017faster} & 0 & 69.97 & 0 & 25.17 & 26508 \\  
PROSER~\cite{zhou2021learning} & 28.73 & 67.81 & \underline{15.36} & 26.63 & 23026 \\ 
DS~\cite{miller2018dropout} & 16.97 & 69.70 & 8.64 & 23.15 & 20108 \\  
OpenDet~\cite{han2022expanding} & \underline{30.98} & \bf{71.29} & 14.84 & \bf{17.12} & \underline{15309} \\
\bf{UADet} & \bf{51.61} & \underline{70.99} & \bf{15.91} & \underline{17.58} & \bf{14347} \\
\midrule
\end{tabular}%
}
\label{tab:Comparisons_with_other methods_on_VOC-COCO-10000}
\end{table}

\section{Additional Ablation Studies}
\label{sec:ablation_add}

\subsection{Different combinations of appearance and geometry uncertainty}
We have experimented with various combinations of appearance uncertainty and geometry uncertainty to model the probability of a negative proposal belonging to the ``unknown" class, as shown in Tab.~\ref{tab:diff_choice}. We observe that different combinations of these uncertainty measures yield comparable open-set detection performance, with minimal variations across different configurations. Based on these empirical results, we adopt configuration (e) as our default setting.
\label{sec:add_quan}
\begin{table}[t]
    \caption{{\bf Performance comparisons with other methods on VOC-COCO-20 dataset.} ``App.'' and ``Geo.'' denote appearance and geometry uncertainty, respectively. $\hat{o}$ and $u$ represent the confidence score output by the RPN and the IoU between the proposal and its corresponding GT, respectively. }
    \centering
\small
\begin{tabular}{p{3mm}|c|c|p{5.5mm}cp{7mm}cp{20mm}cc}
\hline
&App.  &Geo. & WI$_\downarrow$ & AOSE$_\downarrow$ & mAP$_\mathcal{K \uparrow}$ & R$_{\mathcal{U \uparrow}}$ \\
\hline
\textbf{(a)} & $\hat{o}^2$ & $(1 - \text{IoU})^{2}$ & 13.22 & 12565 & 62.39 & 63.39 \\
\textbf{(b)} & $\hat{o}^2$ & $1 - \text{IoU}$ & 12.65 & 12965 & 62.33 & 63.66 \\
\textbf{(c)} & $\hat{o}^2$ & $\sqrt{(1 -\text{IoU})}$ & 12.99 & 13038 & 62.67 & 63.49 \\
\textbf{(d)} & $\hat{o}$ & $(1 - \text{IoU})^{2}$ & 12.39 & 12663 & 62.33 & 67.20 \\
\textbf{(e)} & $\hat{o}$ & $1 - \text{IoU}$ & 12.91 & 12315 & 62.73  & 67.61 \\
\textbf{(f)} & $\hat{o}$ & $\sqrt{(1 -\text{IoU})}$ & 13.41 & 12525 & 62.31 & 66.23 \\
\hline
\end{tabular}


\label{tab:diff_choice}
\end{table}

\subsection{Comparison with the top-\textit{k} one-hot hard labeling strategy} 
In order to validate the superiority of UADet over the one-hot hard labeling strategy, we conduct experiments by assigning negative proposals with top-\textit{k} 
(\textit{k}=$1,5,10$) confidence scores one-hot hard labels and our proposed uncertainty-aware soft labels, respectively. 
\begin{table}[htb]
\caption{\textbf{Comparison with top-\textit{k} hard labeling strategies on VOC-COCO-20.}}

\centering
\small
\setlength{\tabcolsep}{4mm}{
\begin{tabular}{c|c|c|c}
\hline
\multicolumn{2}{c|}{Method} & U-AP ($\uparrow$) & U-Recall ($\uparrow$) \\
\hline
\multirow{2}{*}{top-1} & \multirow{1}{*}{hard} & 13.91 & 36.60 \\
\cline{2-4}          
& \multirow{1}{*}{UADet} & \textbf{15.41} & \textbf{39.68} \\
\cline{1-4}
\multirow{2}{*}{top-5} & \multirow{1}{*}{hard} & 12.96 & 41.41 \\
\cline{2-4}          
& \multirow{1}{*}{UADet} & \textbf{16.05} & \textbf{44.97} \\
\cline{1-4}
\multirow{2}{*}{top-10} & \multirow{1}{*}{hard} & 15.02 & 47.23 \\
\cline{2-4}          
& \multirow{1}{*}{UADet} & \textbf{17.14} & \textbf{48.74} \\
\cline{1-4}
\hline
\end{tabular}}

    \label{tab:topk}
\end{table}
The results are presented in Table~\ref{tab:topk}, which reveal the following conclusions: (1) When the $k$ value is the same, UADet and the top-\textit{k} method exhibit similar U-R. (2) At equivalent levels of unknown recall, UADet demonstrates higher U-AP. (3) As $k$ increases, U-AP of the one-hot hard labeling strategy declines, while UADet remains relatively stable. These observations can be attributed to two reasons. 
First, the RPN generates reliable confidence scores that serve as high-quality supervision signals. Second, unlike the one-hot hard labeling approach, UADet employs uncertainty-aware soft labels that prevent overconfident assignments to pure background or partially unknown objects, thereby consistently achieving better open-set performance.

\end{document}